\documentclass{article} %
\usepackage{iclr2027_conference,times}

\usepackage{amsmath,amsfonts,bm}

\def\eqref#1{equation~\ref{#1}}

\def\1{\bm{1}}

\DeclareMathAlphabet{\mathsfit}{\encodingdefault}{\sfdefault}{m}{sl}
\SetMathAlphabet{\mathsfit}{bold}{\encodingdefault}{\sfdefault}{bx}{n}

\usepackage{hyperref}
\usepackage{url}

\usepackage{bbm}
\usepackage{enumitem}
\usepackage{amssymb}
\usepackage{booktabs}
\usepackage{tabularx}
\usepackage[table]{xcolor}
\usepackage{array}
\usepackage{caption}
\usepackage{tcolorbox}
\usepackage{amsthm}
\usepackage{bm}
\usepackage{wrapfig}
\usepackage[utf8]{inputenc}
\usepackage{hyperref}
\usepackage{graphicx}
\usepackage{xcolor}
\usepackage{fontawesome5}
\usepackage{multirow, multicol}
\usepackage[capitalize]{cleveref}
\usepackage{footnote}
\usepackage{tablefootnote}
\usepackage{listings}
\usepackage{textcomp}
\usepackage{needspace}
\makesavenoteenv{tabular}

\definecolor{projblue}{HTML}{2563EB}
\definecolor{softblack}{HTML}{333333}

\newcommand{\METHODNAME}{\textsc{LeanGraphSearch}}

\title{Lean Graph Search for Formal Theorem \\ Proving}

\author{
  Yihan Xu\thanks{\hspace{0.5mm}Equal contribution.}, 
  ~Jia-Chen Gu\footnotemark[1], 
  ~Junyi Zhang,
  ~Di Wu, 
  ~Haikang Deng,
  ~Po-Nien Kung \\
  {\bf 
  ~Kai-Wei Chang,
  ~Amit Sahai,
  ~Wei Wang,
  ~Nanyun Peng
  } \\
  ~University of California, Los Angeles \\
  \texttt{\{simpson561@, gujc@, junyizhang2002@, diwu@cs.\}ucla.edu} \\
  [3pt]
  {\normalsize
  \href{https://PlusLabNLP.github.io/LeanGraphSearch/}
  {\textcolor{projblue}{\faGlobe}\hspace{2pt}Website}
  \quad
  \href{https://github.com/PlusLabNLP/LeanGraphSearch}
  {\textcolor{softblack}{\faGithub}\hspace{2pt}Code}
  }
}

\usepackage{graphicx}
\usepackage{amsmath}
\usepackage{tikz}
\usetikzlibrary{arrows.meta}

\providecommand{\METHODNAME}{\textsc{LeanGraph}}

\newcommand{\LGFigQRBaseline}{82.7}
\newcommand{\LGFigQROurs}{94.7}
\newcommand{\LGFigMPRBaseline}{39.1}
\newcommand{\LGFigMPROurs}{42.8}
\newcommand{\LGFigFateNone}{52}
\newcommand{\LGFigFateBaseline}{73}
\newcommand{\LGFigFateOurs}{81}

\newcommand{\LGCostOffNoRetrieval}{1302.95}
\newcommand{\LGCostOffLSStd}{1525.68}
\newcommand{\LGCostOffGraphStd}{1495.96}
\newcommand{\LGCostOffLSReason}{1989.28}
\newcommand{\LGCostOffGraphReason}{1985.57}

\newcommand{\LGSolvedOffNoRetrieval}{94}
\newcommand{\LGSolvedOffLSStd}{136}
\newcommand{\LGSolvedOffGraphStd}{143}
\newcommand{\LGSolvedOffLSReason}{145}
\newcommand{\LGSolvedOffGraphReason}{143}

\newcommand{\LGCostOnNoRetrieval}{1413.95}
\newcommand{\LGCostOnLSStd}{1551.38}
\newcommand{\LGCostOnGraphStd}{1553.62}
\newcommand{\LGCostOnLSReason}{2021.83}
\newcommand{\LGCostOnGraphReason}{1966.77}

\newcommand{\LGSolvedOnNoRetrieval}{99}
\newcommand{\LGSolvedOnLSStd}{134}
\newcommand{\LGSolvedOnGraphStd}{142}
\newcommand{\LGSolvedOnLSReason}{142}
\newcommand{\LGSolvedOnGraphReason}{151}

\definecolor{LGFigInk}{HTML}{202C38}
\definecolor{LGFigMuted}{HTML}{586572}
\definecolor{LGFigLine}{HTML}{CBD2D8}
\definecolor{LGFigGrid}{HTML}{E6EAED}
\definecolor{LGFigPale}{HTML}{F7F9FA}
\definecolor{LGFigBlue}{HTML}{416FA6}
\definecolor{LGFigBluePale}{HTML}{ECF2FA}
\definecolor{LGFigTeal}{HTML}{157F79}
\definecolor{LGFigTealPale}{HTML}{E9F5F1}
\definecolor{LGFigGray}{HTML}{9AA6B2}
\definecolor{LGFigGrayPale}{HTML}{F0F2F4}
\definecolor{LGFigGold}{HTML}{A97824}
\definecolor{LGFigGoldPale}{HTML}{FAF2DD}

\newcommand{\LGFigName}{{\rmfamily\METHODNAME{}}}
\newcommand{\LGFigsmall}{\fontsize{7}{8.4}\selectfont}
\newcommand{\LGFignormal}{\fontsize{8}{9.6}\selectfont}
\newcommand{\LGFigheading}{\fontsize{9}{10.8}\selectfont\bfseries}
\tikzset{
  lgfig/base/.style={font=\sffamily\LGFignormal,text=LGFigInk},
  lgfig/heading/.style={font=\sffamily\LGFigheading,anchor=west,inner sep=0pt},
  lgfig/text/.style={font=\LGFignormal,align=center,inner sep=1pt},
  lgfig/note/.style={font=\sffamily\LGFigsmall,text=LGFigMuted,align=center,inner sep=1pt},
  lgfig/box/.style={draw=LGFigLine,fill=white,line width=.65pt,rounded corners=2pt,align=center,inner sep=3pt},
  lgfig/seed/.style={draw=LGFigBlue,fill=LGFigBluePale,text=LGFigBlue,line width=.85pt,rounded corners=2pt,minimum height=.48cm,inner sep=2pt},
  lgfig/premise/.style={draw=LGFigTeal,fill=LGFigTealPale,text=LGFigTeal,line width=.85pt,rounded corners=2pt,minimum height=.48cm,inner sep=2pt},
  lgfig/final/.style={draw=LGFigGold,fill=LGFigGoldPale,text=LGFigGold,line width=.85pt,rounded corners=2pt,minimum height=.48cm,inner sep=2pt},
  lgfig/other/.style={circle,draw=LGFigGray,fill=LGFigGrayPale,line width=.65pt,minimum size=.42cm,inner sep=1pt,text=LGFigMuted},
  lgfig/edge/.style={-{Stealth[length=1.55mm,width=1.15mm]},draw=LGFigMuted,line width=.75pt},
  lgfig/dependency/.style={-{Stealth[length=1.65mm,width=1.2mm]},draw=LGFigTeal,line width=1pt},
  lgfig/semantic/.style={-{Stealth[length=1.55mm,width=1.15mm]},draw=LGFigBlue,line width=.8pt,dash pattern=on 3pt off 2pt},
  lgfig/flow/.style={-{Stealth[length=1.65mm,width=1.2mm]},draw=LGFigInk,line width=.8pt},
  lgfig/panel/.style={draw=LGFigLine,fill=LGFigPale,line width=.6pt,rounded corners=3pt},
  lgfig/card/.style={draw=LGFigLine,fill=white,line width=.65pt,rounded corners=2pt,align=center,inner sep=2pt},
  lgfig/input/.style={draw=LGFigLine,fill=white,line width=.6pt,rounded corners=2pt,align=center,inner sep=3pt,font=\sffamily\LGFigsmall}
}

\newcommand{\LeanGraphTeaserFigure}[1][t]{%
\begin{figure}[#1]
\centering
\pgfmathsetmacro{\LGFigQRGain}{\LGFigQROurs-\LGFigQRBaseline}
\pgfmathsetmacro{\LGFigMPRGain}{\LGFigMPROurs-\LGFigMPRBaseline}
\pgfmathsetmacro{\LGFigFateGain}{\LGFigFateOurs-\LGFigFateBaseline}
\resizebox{\linewidth}{!}{%
\begin{tikzpicture}[lgfig/base,x=1cm,y=1cm,text depth=0.25ex]
\path[use as bounding box] (0,-6.90) rectangle (14.0,.16);
\node[lgfig/heading] at (0,0) {(a) Declaration Dependencies};
\node[lgfig/heading] at (6.08,0) {(b) Retrieval and Proving};
\draw[LGFigLine,line width=.6pt] (5.8,-.42)--(5.8,-6.6);

\node[lgfig/box,minimum width=5.05cm,minimum height=.62cm] (query) at (2.75,-.78)
  {Query $q$: target theorem or proof goal};
\draw[lgfig/panel] (.08,-1.43) rectangle (5.52,-5.91);
\node[lgfig/seed,minimum width=1.72cm,minimum height=.59cm] (d1) at (1.34,-2.22) {Related $d_1$};
\node[lgfig/seed,minimum width=1.72cm,minimum height=.59cm] (d2) at (4.17,-2.22) {Related $d_2$};
\draw[lgfig/semantic] (query.south -| d1.north) -- (d1.north);
\draw[lgfig/semantic] (query.south -| d2.north) -- (d2.north);
\node[lgfig/premise,minimum width=2.12cm,minimum height=.61cm] (l1) at (2.12,-3.63) {Shared premise $\ell_1$};
\node[lgfig/premise,minimum width=2.10cm,minimum height=.61cm] (l2) at (3.92,-5.02) {Multi-hop premise $\ell_2$};
\node[lgfig/other] (u1) at (.70,-4.97) {$d_3$};
\node[lgfig/other] (u2) at (4.91,-3.61) {$d_4$};
\draw[lgfig/dependency] (d1.south) -- (l1.north west);
\draw[lgfig/dependency] (d2.south west) -- (l1.north east);
\draw[lgfig/dependency] (l1.south east) -- (l2.north west);
\draw[lgfig/edge,draw=LGFigGray] (d2.south east)--(u2.north);
\draw[lgfig/edge,draw=LGFigGray] (l1.south west)--(u1.north east);

\draw[lgfig/semantic] (.15,-6.18)--(.62,-6.18);
\node[lgfig/note,anchor=west,inner sep=0pt] at (.70,-6.18) {query match};

\draw[lgfig/dependency] (2.78,-6.18)--(3.25,-6.18);
\node[lgfig/note,anchor=west,inner sep=0pt] at (3.32,-6.18) {formal dependency};

\node[
  lgfig/other,
  minimum width=.38cm,
  minimum height=.22cm,
  inner sep=0pt,
  outer sep=0pt
] at (.385,-6.55) {};
\node[lgfig/note,anchor=west,inner sep=0pt] at (.70,-6.55) {other declarations in graph};

\node[lgfig/text,anchor=west,font=\sffamily\LGFignormal\bfseries] at (6.08,-.59)
  {MathlibQR Recall (\%)};
\node[lgfig/note,text=LGFigTeal,anchor=east,font=\sffamily\LGFigsmall\bfseries] at (14.0,-.59)
  {$+\pgfmathprintnumber[fixed,precision=1,zerofill]{\LGFigQRGain}$ pp vs. LeanSearch v2};
\foreach \x/\lab in {8.77/0,11.095/50,13.42/100}{
  \node[lgfig/note,anchor=north] at (\x,-2.03) {\lab};
}
\foreach \x/\lab in {8.77/0}{
  \draw[LGFigGrid,line width=.5pt] (\x,-.92)--(\x,-1.85);
}
\draw[LGFigLine,line width=.6pt] (8.77,-1.86)--(13.42,-1.86);
\foreach \val/\yy/\name/\col in {
  \LGFigQRBaseline/-1.12/LeanSearch v2/LGFigBlue,
  \LGFigQROurs/-1.57/\LGFigName/LGFigTeal}{
  \pgfmathsetmacro{\xx}{8.77+.0465*\val}
  \fill[\col] (8.77,\yy-.145) rectangle (\xx,\yy+.145);
  \node[lgfig/text,anchor=east] at (8.64,\yy) {\name};
  \node[lgfig/text,anchor=west,text=LGFigInk,font=\sffamily\LGFignormal\bfseries] at (\xx+.10,\yy)
    {\pgfmathprintnumber[fixed,precision=1,zerofill,assume math mode=true]{\val}};
}

\node[lgfig/text,anchor=west,font=\sffamily\LGFignormal\bfseries] at (6.08,-2.54)
  {MathlibMPR Covered (\%)};
\node[lgfig/note,text=LGFigTeal,anchor=east,font=\sffamily\LGFigsmall\bfseries] at (14.0,-2.54)
  {$+\pgfmathprintnumber[fixed,precision=1,zerofill]{\LGFigMPRGain}$ pp vs. LeanSearch v2};
\foreach \x/\lab in {8.77/0,11.095/50,13.42/100}{
  \node[lgfig/note,anchor=north] at (\x,-3.98) {\lab};
}
\foreach \x/\lab in {8.77/0}{
  \draw[LGFigGrid,line width=.5pt] (\x,-2.87)--(\x,-3.80);
}
\draw[LGFigLine,line width=.6pt] (8.77,-3.81)--(13.42,-3.81);
\foreach \val/\yy/\name/\col in {
  \LGFigMPRBaseline/-3.07/LeanSearch v2/LGFigBlue,
  \LGFigMPROurs/-3.52/\LGFigName/LGFigTeal}{
  \pgfmathsetmacro{\xx}{8.77+.0465*\val}
  \fill[\col] (8.77,\yy-.145) rectangle (\xx,\yy+.145);
  \node[lgfig/text,anchor=east] at (8.64,\yy) {\name};
  \node[lgfig/text,anchor=west,text=LGFigInk,font=\sffamily\LGFignormal\bfseries] at (\xx+.10,\yy)
    {\pgfmathprintnumber[fixed,precision=1,zerofill,assume math mode=true]{\val}};
}

\node[lgfig/text,anchor=west,font=\sffamily\LGFignormal\bfseries] at (6.08,-4.44)
  {FATE-H Success Rate (\%)};
\node[lgfig/note,text=LGFigTeal,anchor=east,font=\sffamily\LGFigsmall\bfseries] at (14.0,-4.44)
  {$+\pgfmathprintnumber[fixed,precision=1,zerofill]{\LGFigFateGain}$ pp vs. LeanSearch v2};
\foreach \x/\lab in {8.77/0,11.095/50,13.42/100}{
  \node[lgfig/note,anchor=north] at (\x,-6.41) {\lab};
}
\foreach \x/\lab in {8.77/0}{
  \draw[LGFigGrid,line width=.5pt] (\x,-4.79)--(\x,-6.23);
}
\draw[LGFigLine,line width=.6pt] (8.77,-6.24)--(13.42,-6.24);
\foreach \val/\yy/\name/\col in {
  \LGFigFateNone/-5.04/No retrieval/LGFigGray,
  \LGFigFateBaseline/-5.49/LeanSearch v2/LGFigBlue,
  \LGFigFateOurs/-5.94/\LGFigName/LGFigTeal}{
  \pgfmathsetmacro{\xx}{8.77+.0465*\val}
  \fill[\col] (8.77,\yy-.145) rectangle (\xx,\yy+.145);
  \node[lgfig/text,anchor=east] at (8.64,\yy) {\name};
  \node[lgfig/text,anchor=west,text=LGFigInk,font=\sffamily\LGFignormal\bfseries] at (\xx+.10,\yy)
    {\pgfmathprintnumber[fixed,precision=0,assume math mode=true]{\val}};
}
\end{tikzpicture}
}%
\vspace{-2mm}
\caption{
\textbf{Dependency-guided retrieval with \METHODNAME{}.}
\textbf{(a)} An illustration of the dependency graph, showing how semantically retrieved declarations (blue) lead to supporting premises (green) through shared and multi-hop dependencies.
\textbf{(b)} Selected retrieval and proving results. Gains are reported in percentage points relative to LeanSearch~v2. 
}
\vspace{-2mm}
\label{fig:teaser}
\end{figure}
}%

\newcommand{\LeanGraphMethodFigure}[1][t]{%
\begin{figure}[#1]
\centering
\resizebox{\linewidth}{!}{%
\begin{tikzpicture}[lgfig/base,x=1cm,y=1cm,text depth=0.25ex]
\path[use as bounding box] (0,-8.71) rectangle (14.0,.16);
\node[lgfig/heading] at (0,0) {(a) Dependency Graph Construction};

\draw[lgfig/box,fill=LGFigGrayPale] (.18,-.49) rectangle (3.53,-2.26);
\draw[lgfig/box] (.06,-.61) rectangle (3.41,-2.38);
\node[lgfig/text,anchor=west,font=\sffamily\LGFignormal\bfseries] at (.23,-.88) {Mathlib Declarations};
\draw[LGFigLine,line width=.45pt] (.25,-1.09)--(3.21,-1.09);
\node[lgfig/text,anchor=west] at (.28,-1.35) {declaration $d_i$};
\node[lgfig/text,anchor=west] at (.28,-1.71) {type: $\cdots\,d_j\,\cdots$};
\node[lgfig/text,anchor=west] at (.28,-2.07) {value: $\cdots\,d_k\,\cdots$};
\draw[lgfig/flow] (3.54,-1.49)--(4.13,-1.49);

\node[lgfig/box,minimum width=2.57cm,minimum height=1.29cm] (jixia) at (5.52,-1.49)
  {\texttt{\bfseries jixia}\\[3pt]Static Analysis\\[2pt]{\LGFigsmall Reference Extraction}};
\draw[lgfig/flow] (6.88,-1.49)--(7.48,-1.49);

\draw[lgfig/panel] (7.60,-.61) rectangle (13.94,-2.38);
\node[lgfig/text,anchor=west,font=\sffamily\LGFignormal\bfseries] at (7.82,-.89)
  {Declaration Dependency Graph $\mathcal{G}=(V,E)$};
\node[lgfig/other,minimum size=.52cm] (gi) at (8.55,-1.62) {$d_i$};
\node[lgfig/other,minimum size=.52cm] (gj) at (10.46,-1.33) {$d_j$};
\node[lgfig/other,minimum size=.52cm] (gk) at (10.46,-2.00) {$d_k$};
\node[lgfig/other,minimum size=.52cm] (gl) at (12.61,-1.66) {$d_\ell$};
\draw[lgfig/edge] (gi.north east)--(gj.west);
\draw[lgfig/edge] (gi.south east)--(gk.west);
\draw[lgfig/edge] (gj.east)--(gl.north west);
\draw[lgfig/edge] (gk.east)--(gl.south west);
\node[lgfig/note,fill=LGFigPale] at (9.41,-1.20) {type};
\node[lgfig/note,fill=LGFigPale] at (9.36,-2.09) {value};
\node[lgfig/note,anchor=west,align=left] at (.05,-2.71)
  {Node: declaration statement and description};
\node[lgfig/note,anchor=east,align=right] at (13.94,-2.71)
  {Edge: declaration $\longrightarrow$ referenced dependency};
\draw[LGFigLine,line width=.6pt] (0,-3.03)--(14,-3.03);

\node[lgfig/heading] at (0,-3.34) {(b) Graph-based Retrieval};

\node[lgfig/input,minimum width=3.55cm,minimum height=.44cm] (qin) at (2.04,-3.92)
  {Query $q$};
\node[lgfig/input,minimum width=3.50cm,minimum height=.44cm] (gin) at (6.91,-3.92)
  {Dependency Graph $\mathcal{G}$};
\node[lgfig/input,minimum width=3.55cm,minimum height=.44cm] (qagain) at (11.84,-3.92)
  {Query $q$};

\draw[lgfig/panel] (.03,-4.42) rectangle (4.07,-7.86);
\draw[lgfig/panel] (4.79,-4.42) rectangle (9.03,-7.86);
\draw[lgfig/panel] (9.75,-4.42) rectangle (13.97,-7.86);
\draw[lgfig/flow] (qin.south)--(2.04,-4.40);
\draw[lgfig/flow] (gin.south)--(6.91,-4.40);
\draw[lgfig/flow] (qagain.south)--(11.84,-4.40);
\node[lgfig/text,font=\sffamily\LGFignormal\bfseries] at (2.05,-4.69) {1. Semantic Seed Retrieval};
\node[lgfig/text,font=\sffamily\LGFignormal\bfseries] at (6.91,-4.69) {2. Query-Aware Propagation};
\node[lgfig/text,font=\sffamily\LGFignormal\bfseries] at (11.86,-4.69) {3. Candidate Refinement};

\node[lgfig/card,minimum width=3.48cm,minimum height=.68cm] (dense) at (2.05,-5.38)
  {Embedding Retrieval};
\node[lgfig/card,minimum width=3.48cm,minimum height=.68cm] (rerank1) at (2.05,-6.39)
  {First Reranking};
\node[lgfig/seed,minimum width=3.48cm,minimum height=.58cm,align=center] (seeds) at (2.05,-7.37)
  {Top $K$ Declarations Set $S$};
\draw[lgfig/flow] (dense.south)--(rerank1.north);
\draw[lgfig/flow] (rerank1.south)--(seeds.north);
\draw[lgfig/flow] (4.10,-6.27)--(4.76,-6.27);
\node[lgfig/note,fill=white] at (4.43,-5.98) {$S, q$};

\node[lgfig/note,text=LGFigTeal,font=\sffamily\LGFigsmall\bfseries] at (6.91,-5.10) {Personalized PageRank};
\draw[draw=LGFigBlue!35,fill=LGFigBluePale,rounded corners=3pt,line width=.55pt]
  (5.07,-5.51) rectangle (6.00,-6.63);
\node[lgfig/seed,minimum width=.46cm,minimum height=.40cm,inner sep=1pt] (s1) at (5.53,-5.77) {$d_1$};
\node[lgfig/seed,minimum width=.46cm,minimum height=.40cm,inner sep=1pt] (s2) at (5.53,-6.37) {$d_2$};
\node[lgfig/premise,minimum width=.47cm,minimum height=.44cm,inner sep=1pt] (p1) at (6.94,-6.01) {$\ell_1$};
\node[lgfig/premise,minimum width=.47cm,minimum height=.44cm,inner sep=1pt] (p2) at (8.25,-6.48) {$\ell_2$};
\node[lgfig/other,minimum size=.38cm,font=\LGFigsmall] (p3) at (8.29,-5.56) {$d_3$};
\draw[lgfig/dependency] (s1.east)--(p1.north west);
\draw[lgfig/dependency] (s2.east)--(p1.south west);
\draw[lgfig/dependency] (p1.south east)--(p2.west);
\draw[lgfig/edge,draw=LGFigGray] (s1.north east) to[out=20,in=165] (p3.west);
\draw[lgfig/semantic,rounded corners=3pt] (p2.south) -- (8.25,-6.87) -- (5.53,-6.87) -- (s2.south);
\node[lgfig/note,fill=LGFigPale,inner sep=1.5pt] at (6.93,-6.87) {restart};
\node[lgfig/premise,minimum width=3.48cm,minimum height=.48cm,align=center] at (6.92,-7.37)
  {PPR Score Top $N$ Set $T$};
\draw[lgfig/flow] (9.06,-6.27)--(9.72,-6.27);
\node[lgfig/note,fill=white] at (9.39,-5.98) {$S,T$};

\node[lgfig/card,minimum width=3.66cm,minimum height=.68cm] (merge) at (11.86,-5.38)
  {Merge Candidates};
\node[lgfig/card,minimum width=3.66cm,minimum height=.68cm] (rerank2) at (11.86,-6.39)
  {Second Reranking};
\node[lgfig/card,minimum width=3.66cm,minimum height=.58cm,align=center] (mix) at (11.86,-7.37)
  {Combine Scores};
\draw[lgfig/flow] (merge.south)--(rerank2.north);
\draw[lgfig/flow] (rerank2.south)--(mix.north);
\node[lgfig/final,minimum width=4.18cm,minimum height=.43cm,align=center] (out) at (11.86,-8.45)
  {Final Top $K$ Declarations};
\draw[lgfig/flow] (mix.south)--(out.north);

\node[lgfig/seed,minimum width=.26cm,minimum height=.22cm,inner sep=0pt] at (.17,-8.44) {};
\node[lgfig/note,anchor=west] at (0.42,-8.44) {Semantic Seeds};
\node[lgfig/premise,minimum width=.26cm,minimum height=.22cm,inner sep=0pt] at (2.78,-8.44) {};
\node[lgfig/note,anchor=west] at (3.03,-8.44) {Graph-Expanded Candidates};
\end{tikzpicture}
}%
    \caption{
    \textbf{Overview of \METHODNAME{}.}
    \textbf{(a)} Dependency graph construction (\cref{sec:method:graph}).
    Static analysis with \texttt{jixia} extracts references from Mathlib declarations to construct a directed graph $\mathcal{G}$.
    \textbf{(b)} Graph-based retrieval (\cref{sec:method:retrieval}).
    It first retrieves semantic seeds using semantic embeddings, then expands the candidate set by traversing the dependency graph. Seed and graph-expanded candidates are merged and reranked to select the final top-$K$ declarations.
    \vspace{-3mm}
    }
    \label{fig:method}
\end{figure}
}%

\newcommand{\LGDrawCostBarCompact}[5]{%
  \pgfmathsetmacro{\LGAxisStart}{#1 + 1.85}%
  \pgfmathsetmacro{\LGAxisScale}{4.05/16.0}%
  \pgfmathsetmacro{\LGBarEndX}{\LGAxisStart + \LGAxisScale*(#2)}%
  \pgfmathsetmacro{\LGLabelX}{#1 + 1.68}%
  \fill[#5] (\LGAxisStart,#3-.15) rectangle (\LGBarEndX,#3+.15);
  \node[lgfig/text,anchor=east] at (\LGLabelX,#3) {#4};
  \node[
    lgfig/text,
    anchor=west,
    text=LGFigInk,
    font=\sffamily\LGFignormal\bfseries
  ] at (\LGBarEndX+.08,#3)
    {\$\pgfmathprintnumber[fixed,precision=2]{#2}};
}

\newcommand{\LeanGraphCostEfficiencyFigure}[1][t]{%
\begin{figure}[#1]
\centering
\pgfmathsetmacro{\LGRatioOffNoRetrieval}{\LGCostOffNoRetrieval/\LGSolvedOffNoRetrieval}
\pgfmathsetmacro{\LGRatioOffLSStd}{\LGCostOffLSStd/\LGSolvedOffLSStd}
\pgfmathsetmacro{\LGRatioOffGraphStd}{\LGCostOffGraphStd/\LGSolvedOffGraphStd}
\pgfmathsetmacro{\LGRatioOffLSReason}{\LGCostOffLSReason/\LGSolvedOffLSReason}
\pgfmathsetmacro{\LGRatioOffGraphReason}{\LGCostOffGraphReason/\LGSolvedOffGraphReason}

\pgfmathsetmacro{\LGRatioOnNoRetrieval}{\LGCostOnNoRetrieval/\LGSolvedOnNoRetrieval}
\pgfmathsetmacro{\LGRatioOnLSStd}{\LGCostOnLSStd/\LGSolvedOnLSStd}
\pgfmathsetmacro{\LGRatioOnGraphStd}{\LGCostOnGraphStd/\LGSolvedOnGraphStd}
\pgfmathsetmacro{\LGRatioOnLSReason}{\LGCostOnLSReason/\LGSolvedOnLSReason}
\pgfmathsetmacro{\LGRatioOnGraphReason}{\LGCostOnGraphReason/\LGSolvedOnGraphReason}

\resizebox{\linewidth}{!}{%
\begin{tikzpicture}[lgfig/base,x=1cm,y=1cm,text depth=0.25ex]
\path[use as bounding box] (0,-4.55) rectangle (14.0,.12);

\draw[lgfig/panel] (.05,-.18) rectangle (6.92,-4.02);
\draw[lgfig/panel] (7.08,-.18) rectangle (13.95,-4.02);

\node[lgfig/text,anchor=west,font=\sffamily\LGFignormal\bfseries] at (.30,-.55)
  {Failure Memory Disabled};
\node[lgfig/text,anchor=west,font=\sffamily\LGFignormal\bfseries] at (7.33,-.55)
  {Failure memory Enabled};

\draw[LGFigLine,line width=.6pt] (1.90,-3.62)--(5.95,-3.62);
\foreach \dx/\lab in {0/0,1.0125/4,2.025/8,3.0375/12,4.05/16}{
  \draw[LGFigLine,line width=.55pt] ({1.90+\dx},-3.62)--({1.90+\dx},-3.54);
  \node[lgfig/note,anchor=north] at ({1.90+\dx},-3.70) {\lab};
}

\LGDrawCostBarCompact{.05}{\LGRatioOffNoRetrieval}{-1.10}{No retrieval}{LGFigGray}
\LGDrawCostBarCompact{.05}{\LGRatioOffLSStd}{-1.60}{LS Std.}{LGFigBlue}
\LGDrawCostBarCompact{.05}{\LGRatioOffGraphStd}{-2.10}{LGS Std.}{LGFigTeal}
\LGDrawCostBarCompact{.05}{\LGRatioOffLSReason}{-2.60}{LS Reas.}{LGFigBlue}
\LGDrawCostBarCompact{.05}{\LGRatioOffGraphReason}{-3.10}{LGS Reas.}{LGFigTeal}

\draw[LGFigLine,line width=.6pt] (8.93,-3.62)--(12.98,-3.62);
\foreach \dx/\lab in {0/0,1.0125/4,2.025/8,3.0375/12,4.05/16}{
  \draw[LGFigLine,line width=.55pt] ({8.93+\dx},-3.62)--({8.93+\dx},-3.54);
  \node[lgfig/note,anchor=north] at ({8.93+\dx},-3.70) {\lab};
}

\LGDrawCostBarCompact{7.08}{\LGRatioOnNoRetrieval}{-1.10}{No retrieval}{LGFigGray}
\LGDrawCostBarCompact{7.08}{\LGRatioOnLSStd}{-1.60}{LS Std.}{LGFigBlue}
\LGDrawCostBarCompact{7.08}{\LGRatioOnGraphStd}{-2.10}{LGS Std.}{LGFigTeal}
\LGDrawCostBarCompact{7.08}{\LGRatioOnLSReason}{-2.60}{LS Reas.}{LGFigBlue}
\LGDrawCostBarCompact{7.08}{\LGRatioOnGraphReason}{-3.10}{LGS Reas.}{LGFigTeal}

\end{tikzpicture}
}%
\caption{
\textbf{Cost efficiency on theorem proving.}
\textbf{LS} = LeanSearch, \textbf{LGS} = \METHODNAME{}, \textbf{Std.} = Standard Mode, \textbf{Reas.} = Reasoning Mode.
For each of the 10 configurations, we compute total API cost divided by the number of solved problems on the union of 210 theorem-proving instances.
Solved counts are measured by pass@32.
Lower values indicate better cost efficiency.
\vspace{-3mm}
}
\label{fig:cost_efficiency}
\end{figure}
}%

\iclrfinalcopy %
\begin{document}

\maketitle

\begin{abstract}
Retrieving relevant lemmas is crucial for formal theorem proving, enabling provers to build machine-checkable proofs from established mathematical results.
Recent retrieval systems match library declarations to a target theorem or its intermediate proof steps, but make limited use of multi-hop dependencies among declarations.
These dependencies can reveal supporting lemmas that are not necessarily lexically similar to the target theorem but recur in proofs of related problems. %
To this end, we propose \METHODNAME{}, a graph-based retrieval method for Lean~4 that constructs a graph from Mathlib's forward dependencies and traverses it to uncover supporting lemmas through multi-hop dependency chains.
Specifically, \METHODNAME{} links each declaration to the premises it depends on, uses a graph-based ranking algorithm to propagate relevance from semantically matched declarations to their dependencies, and reranks the expanded candidates against the original query.
Across two mathematical retrieval benchmarks, \METHODNAME{} improves recall by up to 12 percentage points over the strongest baseline, LeanSearch~v2, on MathlibQR and increases coverage of complete annotated proof routes on MathlibMPR by up to 3.7 percentage points.
More importantly, we evaluate whether these retrieval gains translate into improved end-to-end theorem proving by integrating \METHODNAME{} into an iterative proving loop guided by Lean feedback.
Across three theorem proving benchmarks, \METHODNAME{} improves pass rate at $\{ 8, 16, 32 \}$ reflection turns by as much as 40\% compared to the vanilla prover and at most 10\% compared to LeanSearch~v2, showing a better accuracy–cost tradeoff.
These findings highlight the value of dependency graph search for theorem proving and motivate further research on structure-aware search for proof discovery.
\end{abstract}
\section{Introduction}
\label{sec:introduction}

Formal theorem proving expresses mathematical statements and proofs in a formal language, allowing a proof assistant such as Lean to verify their correctness~\citep{lean4}. 
Search engines support the proving process by helping users and automated provers locate useful definitions, theorems, and lemmas within large mathematical libraries such as Mathlib~\citep{mathlib,leandojo,magnushammer,realprover,matlas}. 
Existing retrieval systems, such as LeanExplore~\citep{leanexplore}, LeanFinder~\citep{leanfinder}, and LeanSearch~v2~\citep{leansearchv2}, address this need by identifying library declarations relevant to a target theorem or an intermediate goal. 
However, their limited use of multi-hop dependencies among declarations may miss supporting lemmas that share little lexical similarity with the query but occur in proofs of related problems~\citep{realprover,leansearchv2}.

A key observation is that declarations relevant to a target theorem can serve as starting points for discovering supporting lemmas through multi-hop dependencies. 
These dependencies connect declarations to their direct and indirect premises, providing extra paths to useful lemmas that are less lexically similar to the query. %
To capture and exploit this structure, we introduce \textbf{\METHODNAME{}} (\cref{fig:teaser}), a graph-based retrieval framework for Lean~4 theorem proving.
It constructs a directed graph from Mathlib's forward dependencies, representing declarations as nodes and linking each declaration to the premises it directly depends on.
Searching this graph from semantically relevant declarations extends retrieval beyond direct query matching, helping users and automated provers discover supporting premises that may otherwise be difficult to locate.

\LeanGraphTeaserFigure

To search the constructed dependency graph, \METHODNAME{} uses the target query to guide traversal along formal dependencies.
Given an informal description or a formal goal, it first retrieves declarations using embeddings and reranks them against the query to select semantically relevant seeds.
The seed relevance scores are normalized to form the personalization distribution for personalized PageRank~\citep{pagerank}, which propagates relevance along dependency edges while repeatedly returning to the seeds, thereby expanding the candidate set.
Finally, \METHODNAME{} reranks the expanded candidates against the original query and combines these second-pass relevance scores with query-specific graph scores to produce the final ranking.

We evaluate \METHODNAME{} on both standalone premise retrieval and downstream formal theorem proving. 
For premise retrieval, \METHODNAME{} improves Recall@100 from 82.7\% with LeanSearch~v2 to 94.7\% on the 810-query fair subset of MathlibQR~\citep{leansearchv2}, confirming that dependency graph search effectively recovers additional relevant lemmas. 
On MathlibMPR~\citep{leansearchv2}, it improves proof route-level premise coverage, increasing Covered@20 from 39.1\% to 42.8\% and Covered@50 from 41.3\% to 44.2\%. These results suggest that dependency-aware retrieval better recovers complete premise sets along annotated proof routes. %
For formal theorem proving, we integrate \METHODNAME{} into an iterative proving loop equipped with Lean's diagnostic feedback and problem-local failure memory. \METHODNAME{} consistently outperforms the state-of-the-art retrieval baselines across FATE-H~\citep{fate}, MathlibMPR-Prop~\citep{leansearchv2}, and Lean-IMO-Bench~\citep{kung2026leap}, demonstrating that premise retrieval improvements translate into higher formal proof success rates.

\section{Related Work}
\label{sec:related}

Mathematical theorem proving can be carried out informally in natural language and mathematical notation or formally in a language checked by a proof assistant such as Lean~\citep{lean4}. 
Large language model (LLM)-based approaches to formal proving include generating tactics within proof-state search~\citep{leandojo,copra} and generating and refining proof scripts through verifier feedback, often guided by informal proof sketches~\citep{kung2026leap}.
Throughout this paper, mathematical search refers to retrieving library declarations for formal proving, using either informal or formal queries.
For additional related work, see \cref{appd:related}.

\paragraph{Search for Formal Mathematics.}
Prior work shows that retrieving reusable library facts improves formal proof construction~\citep{magnushammer,leandojo,realprover}.
Magnushammer learns premise selection through contrastive training and reranking for Isabelle~\citep{magnushammer}, while ReProver and REAL-Prover condition tactic generation on retrieved Lean premises~\citep{leandojo,realprover}.
LeanPremise supports premise selection in user-specific contexts~\citep{leanpremise}, and LeanStateSearch adapts tokenization and reranking to formal proof-state queries~\citep{leanstatesearch}.
Complementary search systems help locate declarations from mathematical descriptions: LeanExplore combines formal statements with informal descriptions for semantic search~\citep{leanexplore}, and Lean Finder learns intent-aware embeddings from synthesized queries and preference feedback~\citep{leanfinder}.
LeanSearch~v2 extends this line of work to global premise retrieval, combining embedding-based search and reranking with iterative sketch--retrieve--reflect cycles to gather supporting premises for an entire proof~\citep{leansearchv2}.

\paragraph{Graph-based Retrieval.}
Graph structure has been used to guide retrieval in both natural-language corpora and formal libraries.
GraphRAG builds an entity graph and community summaries for query-focused summarization~\citep{graphrag}, while HippoRAG applies personalized PageRank to a knowledge graph for multi-hop retrieval~\citep{hipporag}.
KG$^2$RAG starts from semantically retrieved chunks and uses a knowledge graph to expand and organize evidence~\citep{kg2rag}.
In formal mathematical search, LeanExplore uses declaration dependencies to generate informal descriptions and incorporates PageRank centrality as a ranking feature~\citep{leanexplore}.
LeanSearch~v2 also processes dependencies bottom-up when constructing declaration descriptions~\citep{leansearchv2}.
To learn graph-aware representations for premise selection, \citet{graphpremise} combine textual embeddings with a relational graph neural network over state--premise and premise--premise dependencies.

\section{\METHODNAME{}}
\label{sec:method}

\LeanGraphMethodFigure

We introduce a \textbf{graph-based retrieval framework} for Lean theorem proving that exploits structural relationships among declarations. It represents Mathlib as a directed declaration dependency graph (\cref{fig:method} (a)) and uses query-aware propagation to expand from semantically relevant declarations to their formal prerequisites (\cref{fig:method} (b)).
Semantic relevance and dependency structure jointly guide promising declaration selection, followed by reranking to produce the final ranked list.

\subsection{Task Formulation}
\label{sec:method:formulation}

\paragraph{Mathematical Retrieval.}
We study library declaration retrieval for theorem proving in Lean~4 \citep{lean4}. Let $\mathcal{L} = \{d_i\}_{i=1}^{n}$ denote the Mathlib declaration corpus. Each declaration $d_i$ is identified by its name and associated with its formal statement and, when available, a natural-language description\footnote{Descriptions are taken from the Gemini-generated corpus provided by \citet{leansearchv2}.}. A retrieval query $q$ can contain an informal mathematical description, a formal Lean statement, a proof state, or a combination of these. It may describe a target theorem or request a definition or lemma needed at an intermediate proof step.

Given a query $q$, the retriever returns up to $K$ declarations from $\mathcal{L}$ ranked by retrieval score, denoted $\mathcal{R}(q) = \{d_j\}_{j=1}^{K}$.
We consider two complementary objectives: locating an individual declaration that matches the query and retrieving supporting premises for a target theorem. In the latter setting, a useful declaration may address an intermediate object or a side condition, and several declarations may be needed to support a complete proof. Retrieval must therefore capture the relevance of declarations in a proof, even when their statements differ substantially from the query. 

\paragraph{Theorem Proving.}
Each proving problem is specified by a Lean source $x$ containing one target proof hole. At round $t$, a search policy can issue queries using $x$ and, when available, the preceding proof candidate $Y_{t-1}$ and Lean compiler diagnostic $e_{t-1}$. The retrieved declarations are aggregated into a context $R_t$, which the prover uses to construct or revise the candidate $Y_t$. A successful candidate elaborates without a live \texttt{sorry} and satisfies the proof-hole integrity contract. This separates the query-level retrieval interface from the policy that determines when to invoke it.

\subsection{Declaration Dependency Graph Construction}
\label{sec:method:graph}

\METHODNAME{} constructs a directed graph $\mathcal{G}=(\mathcal{V},\mathcal{E})$ from the structural references in
Mathlib through static analysis with the jixia tool \citep{jixia2024}. Each node $i\in\mathcal{V}$ represents one declaration together with its formal statement and, when available, its natural-language description. A directed edge $i\rightarrow j$ indicates that declaration $i$ references declaration $j$ in its type or value. 
Edges therefore point from declarations to their dependencies, allowing traversal from a query-relevant declaration to the definitions and lemmas underlying its statement or implementation. %

This graph provides explicit formal dependency structure that complements embedding-based semantic similarity. %
Starting from a semantically relevant declaration, traversal can reveal potentially useful supporting declarations, including dependencies shared by declarations with different surface forms. The graph thus augments retrieval with formal structural relationships while leaving the underlying Mathlib corpus unchanged.

\subsection{Graph-based Retrieval}
\label{sec:method:retrieval}

\paragraph{Semantic Seed Retrieval.}
Given a query $q$, \METHODNAME{} first retrieves semantically related declarations as seeds for graph propagation using the embedding--reranker retrieval pipeline~\citep{leansearchv2}. The query is encoded with \texttt{Qwen3-Embedding-8B} \citep{qwen3embedding} to retrieve an initial candidate set. Each query--declaration pair is then scored for relevance using the reranker logits produced by \texttt{Qwen3-Reranker-8B} \citep{qwen3embedding}. The top $K$ declarations ranked by these scores form the semantic seed set $S$.

\paragraph{Query-personalized Propagation.}
To exploit the dependency structure described in \cref{sec:method:graph}, \METHODNAME{} propagates a query-specific seed distribution over the graph. %
Specifically, \METHODNAME{} normalizes the semantic seed scores to obtain a seed distribution $v$ over graph nodes, assigning zero mass to nodes outside the seed set $S$:
\begin{equation}
  v_i = \frac{q_i \cdot \mathbbm{1}[i\in S]}{\sum_{j\in S}q_j}.
  \label{eq:seed-distribution}
\end{equation}
Let $P$ denote the row-stochastic transition matrix of $\mathcal{G}$ and $\mathcal{D}$ the set of dangling nodes. \METHODNAME{} computes personalized PageRank (PPR) \citep{pagerank} by iterating:
\begin{equation}
  p^{(k+1)} = r v + (1-r)p^{(k)}P
  + (1-r)\!\left(\sum_{i\in\mathcal{D}}p_i^{(k)}\right)v,
  \label{eq:ppr}
\end{equation}
where $p^{(0)} = v$. The restart term $rv$ keeps propagation anchored to the seed distribution, while the transition term $(1-r)p^{(k)}P$ expands each seed toward its dependencies. Returning dangling mass to $v$ preserves the query-specific distribution.

\paragraph{Graph-aware Candidate Refinement.}
After PPR propagation, \METHODNAME{} merges the semantic seed set $S$ with the top-$N$ graph-expanded candidates ranked by PPR score.
The retained candidates are then scored against the original query through the Qwen reranker in a second reranking stage.
The final score mixes the second-stage relevance score $\tilde q_i$ with the normalized PPR score weighted by a coefficient $\lambda$:
\begin{equation}
  f_i = \tilde q_i
  + \lambda\,\operatorname{zscore}_{\mathrm{query}}(p_i).
  \label{eq:rank-fusion}
\end{equation}
This procedure evaluates graph-expanded and seed candidates using the same relevance model, with the graph signal helping distinguish candidates with similar reranker scores. Finally, \METHODNAME{} returns the top-$K$ candidates with highest $f_{i}$ as $\mathcal{R}(q)$.

\subsection{Search Loop Policy}
\label{sec:method:variant}

To align with the search loop settings in LeanSearch v2, we evaluate two retrieval-policy variants of \METHODNAME{} that share the same retrieval mechanism but differ in when and how it is invoked.
\textbf{Standard Mode} invokes graph-based retrieval after a failed proof attempt using queries conditioned on the current proof state.
\textbf{Reasoning Mode} additionally uses an LLM to decompose the theorem into queries targeting definitions, high-level proof strategies, and individual proof steps before the first proof attempt. Retrieved evidence is aggregated across subqueries and truncated to $K$ declarations. Readers can refer to \citet{leansearchv2} for more details of the search policies.

\section{Experiments on Mathematical Retrieval}
\label{sec:exp:retrieval}

\subsection{Setup}
\label{sec:exp:retrieval_setting}

\paragraph{Benchmarks and Metrics.} 
We evaluate \METHODNAME{} on \textbf{MathlibQR} and \textbf{MathlibMPR}, following the benchmark definitions and metrics of LeanSearch v2 \citep{leansearchv2}. MathlibQR contains six kinds of natural-language queries, each paired with a single target declaration. We use its fair subset of 810 queries over 171 target declarations shared by all compared systems, and report nDCG@$k$ for $k\in\{1,5,10\}$ and Recall@$k$ for $k\in\{10,50,100\}$. MathlibMPR contains 69 research-level Mathlib theorems with grouped premises and alternative proof routings. Recall@$k$ (group) measures the fraction of premise groups hit by the top-$k$ results, while Covered@$k$ measures whether the retrieved set covers every premise group in at least one complete proof routing. We report both metrics for $k\in\{5,10,20,30,50\}$.

\paragraph{Baselines.} 
For non-agentic search, we compare \METHODNAME{} with three prior embedding-based search methods, \textbf{LeanExplore}, \textbf{LeanFinder} and \textbf{LeanSearch v2 (standard mode)} on MathlibQR \citep{leanexplore,leanfinder,leansearchv2}. 
For agentic search, we compare \METHODNAME{} with three search methods that use informal and formal descriptions as queries, \textbf{INF-X-Retriever}, \textbf{ReasonIR} and \textbf{DIVER} \citep{inf-x-retriever,reasonir,diver}; three search methods that use current proof states as queries, \textbf{ReProver}, \textbf{LeanPremise} and \textbf{LeanStateSearch} \citep{leandojo,leanpremise,leanstatesearch}; and finally with \textbf{LeanSearch v2 (reasoning mode)} that uses a simple concatenated description and status as input. To estimate the robustness across different backbone models, we evaluate LeanSearch v2 (reasoning mode) and \METHODNAME{} (reasoning mode) with Gemini 3.1 Pro and Claude Sonnet 5 as backbone models, then average each metric across the two models.

\subsection{Results}
\label{sec:exp:retrieval_result}

\cref{tab:mathlibqr} shows that \METHODNAME{} consistently improves retrieval performance at every measured cutoff. The gains over LeanSearch v2 increase at larger cutoffs, with nDCG@10 increasing from 0.616 to 0.629, Recall@50 by 7.8 percentage points, and Recall@100 by 12.0 percentage points. 
In \cref{tab:mathlibmpr}, \METHODNAME{} improves group recall at four of the five cutoffs, with the exception of Recall@$10$. Improvements in complete-route coverage are more consistent: Covered@k increases at every cutoff, outperforming LeanSearch v2 by 3.7 percentage points at \(k=20\) and 2.9 percentage points at \(k=50\). %

\begin{table*}[t]%
  \centering
  \caption{Query-to-declaration retrieval on the 810-query fair subset of MathlibQR. LeanExplore and LeanFinder are quoted from \citet{leansearchv2}. The best result in each column is \textbf{bold}.}
  \label{tab:mathlibqr}
  \small
  \setlength{\tabcolsep}{5pt}
  \begin{tabular}{lcccccc}
    \toprule
    & \multicolumn{3}{c}{nDCG@$k$} & \multicolumn{3}{c}{Recall@$k$} \\
    \cmidrule(lr){2-4}\cmidrule(lr){5-7}
    Method & 1 & 5 & 10 & 10 & 50 & 100 \\
    \midrule
    LeanExplore & 0.246 & 0.358 & 0.393 & 0.569 & 0.743 & 0.789 \\
    LeanFinder & 0.370 & 0.514 & 0.533 & 0.698 & 0.824 & 0.875 \\
    LeanSearch v2 (standard mode) & 0.472 & 0.596 & 0.616 & 0.765 & 0.821 & 0.827 \\
    \METHODNAME{} (standard mode) & \textbf{0.473} & \textbf{0.604} & \textbf{0.629} & \textbf{0.794} & \textbf{0.899} & \textbf{0.947} \\
    \bottomrule
  \end{tabular}
\end{table*}

\begin{table*}[t]%
  \centering
  \caption{Global premise retrieval on MathlibMPR. The two evaluated method rows are arithmetic means over Gemini 3.1 Pro and Claude Sonnet 5. Baseline results are quoted from \citet{leansearchv2}. %
  }
  \label{tab:mathlibmpr}
  \small
  \setlength{\tabcolsep}{3.5pt}
  \begin{tabular}{lccccc@{\hspace{6pt}}ccccc}
    \toprule
    \multirow{2}{*}[-0.75ex]{\textbf{Method}} & \multicolumn{5}{c}{Recall@$k$ (group)} & \multicolumn{5}{c}{Covered@$k$} \\
    \cmidrule(lr){2-6}\cmidrule(lr){7-11}
    & 5 & 10 & 20 & 30 & 50 & 5 & 10 & 20 & 30 & 50 \\
    \midrule
    INF-X-Retriever & 21.2 & 28.2 & 33.9 & 40.0 & 42.6 & 15.9 & 18.8 & 21.7 & 24.6 & 24.6 \\
    ReasonIR & 23.9 & 26.9 & 33.3 & 35.6 & 43.8 & 17.4 & 18.8 & 20.3 & 21.7 & 30.4 \\
    DIVER & 27.7 & 38.0 & 46.6 & 52.8 & 55.0 & 18.8 & 24.6 & 29.0 & 33.3 & 37.7 \\
    \midrule
    ReProver & 3.0 & 5.5 & 7.6 & 8.8 & 10.3 & 0.0 & 1.4 & 2.9 & 4.3 & 4.3 \\
    LeanPremise & 4.1 & 4.8 & 8.0 & 9.9 & 11.5 & 1.4 & 1.4 & 4.3 & 4.3 & 5.8 \\
    LeanStateSearch & 6.5 & 9.3 & 12.5 & 16.0 & 20.0 & 2.9 & 2.9 & 7.2 & 10.1 & 13.0 \\
    \midrule
    LeanSearch v2 (reasoning mode) & 36.4 & \textbf{48.2} & 55.5 & 57.8 & 58.2 & 22.5 & 29.7 & 39.1 & 41.3 & 41.3 \\
    \METHODNAME{} (reasoning mode) & \textbf{37.9} & 47.0 & \textbf{56.8} & \textbf{57.9} & \textbf{58.3} & \textbf{23.9} & \textbf{31.9} & \textbf{42.8} & \textbf{43.5} & \textbf{44.2} \\
    \bottomrule
  \end{tabular}
\end{table*}

Overall, these results suggest that graph-based retrieval improves both individual declaration retrieval and the recovery of premise sets that support complete proof routes.
In addition to the quantitative results, our qualitative analysis reveals the following findings:

\vspace{-3mm}
\paragraph{Graph-aware expansion reveals underlying mathematical objects.}
On MathlibQR, semantic retrievers frequently identify constructors, projections, or characteristic theorems whose descriptions closely match the query, but miss the associated parent declarations or deeper dependent declarations because their names and signatures are less lexically similar to the query.
Graph-aware expansion connects these semantic matches to underlying declarations through dependency relationships. 

\vspace{-3mm}
\paragraph{Graph augmentation improves the composition of the retrieved context.}
On MathlibMPR, \METHODNAME{} returns fewer filtered declarations on average than LeanSearch v2, yet retrieves more gold declarations within the top five positions,  improving group recall and complete-route coverage. Route-level examples further illustrate this effect. These results suggest that graph augmentation concentrates relevant premises in its retrieved context instead of merely expanding it. %

Examples illustrating these findings are provided in \cref{appd:case}.

\section{Experiments on Theorem Proving}
\label{sec:exp:proving}

More importantly, we evaluate on formal theorem proving tasks to confirm whether improvements in retrieval quality translate into better proof generation.

\subsection{\METHODNAME{} in an Agentic Proving Loop}
\label{sec:exp:agentic_loop}

To integrate \METHODNAME{} into a verifier-grounded proof loop, we implement it as a search tool that can be called by LLM provers. Each problem instance contains a Lean source $x$ with one target proof hole. At inference round $t$, the prover $M_{\theta}$ parameterized by $\theta$ generates a complete Lean source candidate $Y_t$ conditioned on the problem $x$, the preceding candidate $Y_{t-1}$, the latest Lean compiler diagnostic $e_{t-1}$, optional retrieved declarations $R_t$, and optional failure-memory feedback $m_{t-1}$ as:
\begin{equation}
  Y_t \sim M_{\theta}
  \left(x,Y_{t-1},e_{t-1},R_t,m_{t-1}\right).
  \label{eq:prover-update}
\end{equation}
The Lean compiler executes $Y_t$ and returns either a successful elaboration or a structured diagnostic. Each failed candidate is followed by one revision conditioned on the latest verifier feedback, producing a single candidate lineage.

Building on the common practice of incorporating verifier feedback into iterative proof refinement, we investigate whether \METHODNAME{} can provide useful retrieval feedback that further improves proving performance.
After a failed attempt, the LLM optionally generates five search queries from the problem $x$, current candidate $Y_{t}$, and Lean diagnostic $e_{t}$. Each query invokes \METHODNAME{} and returns $K$ declarations. \METHODNAME{} deduplicates the retrieved declarations and inserts them next to the previous candidate and diagnostic in the revision prompt. A candidate succeeds when the complete Lean source elaborates without a live \texttt{sorry} and satisfies the proof-hole integrity contract.

\paragraph{Optional Formal Failure Memory.}

Our preliminary experiments show that the LLM prover is likely to repeat the same or similar failed proofs across multiple rounds, wasting the inference budget and limiting exploration of alternative proof strategies. 
To mitigate this issue, we introduce a failure memory, which augments the agentic loop with problem-local records of recurrent formal states. For each stable failure, we construct a structured failure state as:
\begin{equation}
  s_t = \left(c_t,\bar e_t,G_t,L_t,\tau_t\right),
  \label{eq:failure-state}
\end{equation}
where $c_t$ is the error category, $\bar e_t$ is the normalized diagnostic, $G_t$ contains unresolved goals, $L_t$ summarizes the local context, and $\tau_t$ is the inferred failing tactic. A deterministic serialization and SHA-256 \citep{sha-256} hash identify repeated states.

When a state recurs, the next prover call receives a compact dead-end record and the failed modifications that led to it, encouraging the prover to explore an alternative repair. The memory also reuses deterministic failures of identical candidates and caches retrieval responses by formal proof state and normalized query. These interventions reduce repeated compilation and retrieval while preserving the single-lineage search structure.

\subsection{Setup}
\label{sec:exp:proving_setting}

\paragraph{Benchmarks and Metrics.}

We consider three benchmarks with diverse fields and complexities: \textbf{FATE-H}~\citep{fate}, \textbf{MathlibMPR-Prop}~\citep{leansearchv2} and \textbf{Lean-IMO-Bench}~\citep{kung2026leap}. FATE-H contains 100 graduate-level algebra problems, MathlibMPR-Prop contains the 50 propositions from MathlibMPR that admit a uniform tactic-mode interface, and Lean-IMO-Bench contains 30 Basic and 30 Advanced Olympiad problems.
For each problem, we allow the backbone model at most 32 reflection turns with one optional query assigned to each turn, and report pass@$k$ for $k \in \{8, 16, 32\}$, where $k$ represents the number of reflection turns. We also report the normalized total cost over each group to assess cost effectiveness. %

\paragraph{Baselines.}

To control evaluation costs while retaining representative baselines, we mostly compare \METHODNAME{} with the state-of-the-art \textbf{LeanSearch v2} \citep{leansearchv2} and the \textbf{vanilla backbone LLM without retrieval}. To align with the searching loop settings in LeanSearch v2, we evaluate both LeanSearch v2 and \METHODNAME{} under the \textbf{Standard Mode} and \textbf{Reasoning Mode}.
Each of the five configurations is evaluated with failure memory both disabled and enabled, respectively. Specifically, each retrieval action produces five queries conditioned on the current theorem, proof candidate, and Lean diagnostic, with each query retrieving up to 50 declarations. Gemini 3.1 Pro is the prover backbone throughout all experiments. Within each benchmark, the prover prompt, verifier, action budget, and model settings are fixed across all configurations.

\subsection{Results}
\label{sec:exp:proving_result}

\cref{tab:proving-fate,tab:proving-mpr,tab:proving-imo} show the pass rates and costs relative to the no-retrieval baseline on the three benchmarks. 
Our results reveal the following findings:

\begin{table*}[t]%
  \centering
  \caption{Proving results on \textbf{FATE-H}, grouped by whether failure memory is enabled. Relative cost is the full-budget API cost normalized by no retrieval within each group. The best and second best results in each group are shown in \textbf{bold} and \underline{underlined}, respectively.}
  \label{tab:proving-fate}
  \small
  \setlength{\tabcolsep}{10pt}
  \begin{tabular}{lcccc}
    \toprule
    \textbf{Retrieval condition} & pass@$8$ & pass@$16$ & pass@$32$ & Relative cost \\
    \midrule
    \rowcolor{gray!15}\multicolumn{5}{l}{\textsc{Failure Memory Disabled}} \\
    No retrieval & 35 & 42 & 47 & $1.00\times$ \\
    LeanSearch v2 (standard mode) & 49 & 61 & 70 & $1.11\times$ \\
    \METHODNAME{} (standard mode) & \underline{53} & \underline{69} & 71 & $1.09\times$ \\
    LeanSearch v2 (reasoning mode) & 52 & 68 & \underline{73} & $1.41\times$ \\
    \METHODNAME{} (reasoning mode) & \textbf{56} & \textbf{70} & \textbf{75} & $1.37\times$ \\
    \midrule
    \rowcolor{gray!15}\multicolumn{5}{l}{\textsc{Failure Memory Enabled}} \\
    No retrieval & 33 & 45 & 52 & $1.00\times$ \\
    LeanSearch v2 (standard mode) & 53 & 63 & 71 & $1.01\times$ \\
    \METHODNAME{} (standard mode) & \underline{54} & \underline{68} & \underline{73} & $1.01\times$ \\
    LeanSearch v2 (reasoning mode) & \underline{54} & 66 & 73 & $1.31\times$ \\
    \METHODNAME{} (reasoning mode) & \textbf{58} & \textbf{75} & \textbf{81} & $1.22\times$ \\
    \bottomrule
  \end{tabular}
\end{table*}

\begin{table*}[t]%
  \centering
  \caption{Proving results on \textbf{MathlibMPR-Prop}, grouped by whether failure memory is enabled. %
  }
  \label{tab:proving-mpr}
  \small
  \setlength{\tabcolsep}{10pt}
  \begin{tabular}{lcccc}
    \toprule
    \textbf{Retrieval condition} & pass@$8$ & pass@$16$ & pass@$32$ & Relative cost \\
    \midrule
    \rowcolor{gray!15}\multicolumn{5}{l}{\textsc{Failure Memory Disabled}} \\
    No retrieval & 15 & 18 & 23 & $1.00\times$ \\
    LeanSearch v2 (standard mode) & 31 & 35 & \underline{39} & $0.82\times$ \\
    \METHODNAME{} (standard mode) & \underline{32} & \textbf{38} & \textbf{41} & $0.78\times$ \\
    LeanSearch v2 (reasoning mode) & \textbf{33} & \underline{37} & \textbf{41} & $1.01\times$ \\
    \METHODNAME{} (reasoning mode) & \textbf{33} & \textbf{38} & \underline{39} & $1.06\times$ \\
    \midrule
    \rowcolor{gray!15}\multicolumn{5}{l}{\textsc{Failure Memory Enabled}} \\
    No retrieval & 15 & 19 & 23 & $1.00\times$ \\
    LeanSearch v2 (standard mode) & \underline{33} & 35 & \underline{39} & $0.76\times$ \\
    \METHODNAME{} (standard mode) & \textbf{35} & \textbf{37} & \textbf{40} & $0.78\times$ \\
    LeanSearch v2 (reasoning mode) & \underline{33} & \underline{36} & \underline{39} & $0.99\times$ \\
    \METHODNAME{} (reasoning mode) & 31 & \textbf{37} & \textbf{40} & $0.96\times$ \\
    \bottomrule
  \end{tabular}
\end{table*}

\begin{table*}[t]%
  \centering
  \caption{Proving results on \textbf{Lean-IMO-Bench}, grouped by whether failure memory is enabled. %
  }
  \label{tab:proving-imo}
  \small
  \setlength{\tabcolsep}{4pt}
  \resizebox{\textwidth}{!}{
  \begin{tabular}{lcccccccc}
    \toprule
    \multirow{2}{*}[-0.75ex]{\textbf{Retrieval condition}} & \multicolumn{4}{c}{Basic Set ($n=30$)} & \multicolumn{4}{c}{Advanced Set ($n=30$)} \\
    \cmidrule(lr){2-5}\cmidrule(lr){6-9}
    & pass@$8$ & pass@$16$ & pass@$32$ & Rel. cost & pass@$8$ & pass@$16$ & pass@$32$ & Rel. cost \\
    \midrule
    \rowcolor{gray!15}\multicolumn{9}{l}{\textsc{Failure Memory Disabled}} \\
    No retrieval & 12 & \underline{16} & 16 & $1.00\times$ & \textbf{7} & 7 & 8 & $1.00\times$ \\
    LeanSearch v2 (standard mode) & \underline{15} & 15 & 18 & $1.49\times$ & 4 & \underline{8} & \underline{9} & $1.82\times$ \\
    \METHODNAME{} (standard mode) & \textbf{16} & \textbf{17} & \textbf{21} & $1.40\times$ & \underline{5} & \textbf{9} & \textbf{10} & $1.88\times$ \\
    LeanSearch v2 (reasoning mode) & 14 & \underline{16} & \textbf{21} & $2.04\times$ & \textbf{7} & \textbf{9} & \textbf{10} & $2.56\times$ \\
    \METHODNAME{} (reasoning mode) & 14 & \underline{16} & \underline{19} & $2.09\times$ & \textbf{7} & \underline{8} & \textbf{10} & $2.53\times$ \\
    \midrule
    \rowcolor{gray!15}\multicolumn{9}{l}{\textsc{Failure Memory Enabled}} \\
    No retrieval & 13 & 15 & 17 & $1.00\times$ & 5 & 6 & 7 & $1.00\times$ \\
    LeanSearch v2 (standard mode) & 13 & 16 & 16 & $1.61\times$ & 6 & 7 & 8 & $1.64\times$ \\
    \METHODNAME{} (standard mode) & \underline{15} & \textbf{18} & \textbf{20} & $1.64\times$ & \textbf{7} & \underline{9} & \underline{9} & $1.61\times$ \\
    LeanSearch v2 (reasoning mode) & \textbf{16} & \textbf{18} & \underline{18} & $1.94\times$ & \textbf{7} & \textbf{10} & \textbf{12} & $2.31\times$ \\
    \METHODNAME{} (reasoning mode) & \underline{15} & \underline{17} & \underline{18} & $2.25\times$ & \textbf{7} & \underline{9} & \textbf{12} & $2.20\times$ \\
    \bottomrule
  \end{tabular}}
\end{table*}

\LeanGraphCostEfficiencyFigure

\paragraph{Graph retrieval improves proof completion.}
\METHODNAME{} in Standard Mode outperforms LeanSearch~v2 in Standard Mode at every evaluated reflection budget across all sets, while achieving performance comparable to LeanSearch~v2 in Reasoning Mode.
The number solved across the 210 problems within 32 turns increases from 136 to 143 without memory and from 134 to 142 with memory.
\METHODNAME{} in Reasoning Mode yields further improvement, reaching the highest total number of solved problems, with particularly strong results on FATE-H.

The benefit is also visible at smaller reflection budgets.
Comparing results across budgets reveals improvements that are less apparent from pass@32 alone.
On FATE-H, Standard Mode without memory solves 69 problems within 16 turns using \METHODNAME{}, compared with 61 using LeanSearch~v2.
In Reasoning Mode with memory, \METHODNAME{} solves 75 FATE-H problems within 16 turns, exceeding the 73 problems solved by LeanSearch~v2 within 32 turns.

\vspace{-3mm}
\paragraph{Failure memory improves proving performance under limited reflection budgets.}
At $k=8$, enabling failure memory increases the total number of solved problems across all four retrieval configurations.
In Standard Mode, the count rises from 106 to 111 for \METHODNAME{} and from 99 to 105 for LeanSearch~v2.
In Reasoning Mode, memory also improves completion, increasing the aggregate pass@16 count for \METHODNAME{} from 132 to 138.
These results show that failure memory can help the prover complete more proofs under limited reflection budgets.

\vspace{-3mm}
\paragraph{Graph retrieval lowers API cost per solved problem.}
\cref{fig:cost_efficiency} reports \emph{API cost per solved problem} for each retrieval configuration across all 210 theorem-proving instances, calculated as total API expenditure, including unsuccessful runs, divided by the number of problems solved within 32 turns. 
\METHODNAME{} in Standard Mode achieves the lowest cost per solved problem.
\METHODNAME{} in Reasoning Mode with memory achieves the highest aggregate completion, solving 151 of 210 problems, while maintaining a lower cost per solved problem than LeanSearch~v2.

Our qualitative analysis suggests that, by following multi-hop dependencies, \METHODNAME{} uncovers supporting lemmas used in related proofs that direct query matching may miss. Retrieving these premises contributes to improved downstream theorem proving. Examples illustrating this mechanism are provided in \cref{appd:case}.

\section{Conclusion}
\label{sec:conclusion}

We introduced \METHODNAME{}, a graph-based retrieval method that exploits formal dependencies in Mathlib to discover supporting premises for Lean~4 theorem proving.
Starting from semantically relevant declarations, it propagates relevance along dependency edges with personalized PageRank and reranks the expanded candidates against the original query.
Experiments show improvements in declaration recall and annotated proof-route coverage, with benefits extending to downstream proof construction.
\METHODNAME{} improves proof completion over matched LeanSearch~v2 configurations on most of the evaluated benchmarks and reflection budgets, while reducing the aggregate API cost per solved problem.
Together, these findings show that the dependencies recorded in a formal library provide useful guidance for retrieving supporting premises, connecting semantic search to the mathematical ingredients needed to construct a formal proof.
\subsection*{AI use statement}
In this work, we used generative AI tools to implement methods and assist with translation. Additionally, we used generative AI tools to create or edit software code, summarize or analyze existing literature, and source or search for information.

We did not use generative AI tools to propose or refine hypotheses, design or provide feedback on research methodology or experiments, clean or reformat datasets, support qualitative or thematic data analysis, interpret results, generate synthetic datasets, help develop theoretical models or conceptual frameworks, formulate mathematical claims, provide critical ingredients for proving mathematical claims, or assist in writing proofs. These uses are not applicable to this work.

All code, writing, and references were double-checked by at least two authors. We take responsibility for the final content of this work, including text, claims, and artifacts produced with the aid of generative AI.

\subsection*{Ethics statement}
This paper mainly discusses formal mathematical declaration search and the corresponding theorem-proving harness. There are several potential societal consequences of our work, none of which we feel must be specifically highlighted here.

\subsection*{Reproducibility statement}
The models and data used in this paper are all accessible online, and we provide citations for all baselines. Readers can refer to the codebases of Gao et al. (2026) and frenzymath (2024) to better reproduce our framework. We have also uploaded a redacted version of our codebase as supplemental material for reference. For more technical details of our implementation setup, see the Appendix.

\bibliography{iclr2027_conference}
\bibliographystyle{iclr2027_conference}

\clearpage
\crefalias{section}{appendix}
\appendix
\section{Related Work}
\label{appd:related}

\subsection{Retrieval Tasks and Benchmarks}
\label{appd:related:retrieval}

General retrieval benchmarks evaluate different aspects of the relationship between queries and supporting evidence.
BEIR measures retrieval across heterogeneous domains~\citep{beir}, while MTEB evaluates embeddings across retrieval and other representation tasks~\citep{mteb}.
BRIGHT emphasizes queries that require reasoning beyond direct semantic matching~\citep{bright}, motivating specialized retrievers such as ReasonIR, INF-X-Retriever, and DIVER~\citep{reasonir,inf-x-retriever,diver}.
In mathematics, NaturalProofs evaluates reference retrieval for natural-language proofs~\citep{naturalproofs}, and Matlas supports statement-level search over mathematical literature~\citep{matlas}.

Formal proof traces provide more explicit supervision for premise use.
HolStep and HOList supply proof-step data from HOL Light~\citep{holstep,holist}, while LeanDojo extracts proof states, tactics, and premise annotations from Lean developments~\citep{leandojo}.
LeanSearch~v2 introduces two complementary retrieval benchmarks~\citep{leansearchv2}: MathlibQR evaluates queries paired with individual target declarations, whereas MathlibMPR annotates premise groups and alternative proof routes.
Their metrics distinguish locating a relevant declaration from covering the supporting premises along a complete route.
Our evaluation combines these retrieval measures with downstream proof completion to examine whether the retrieved declarations help construct checked proofs.

\subsection{Formal Theorem-Proving Systems}
\label{appd:related:systems}

Premise selection has long connected mathematical libraries to automated reasoning.
Sledgehammer and CoqHammer select library facts for external provers and reconstruct successful proofs within a proof assistant~\citep{sledgehammer,coqhammer}.
Thor integrates language models with hammer-based automation~\citep{thor}, and Magnushammer learns premise selection through contrastive training and reranking~\citep{magnushammer}.
Neural provers also explore proof states through learned tactic selection, as in HOList and HyperTree Proof Search~\citep{holist,htps}.
ReProver and REAL-Prover incorporate retrieved Lean premises into this stepwise proving process~\citep{leandojo,realprover}.

Recent work improves the prover through mathematical pretraining~\citep{llemma}, scaffolded data synthesis and self-correction~\citep{goedelprover2}, and reinforcement learning with formal verification~\citep{deepseekprover15,deepseekprover2,alphaproof}.
Informal reasoning provides another source of guidance: Draft, Sketch, and Prove generates formal sketches from informal proofs~\citep{draftsketchprove}, while Lean-STaR, Hilbert, and LEAP connect informal reasoning to formal tactics or subproblems~\citep{leanstar,hilbert,kung2026leap}.
These approaches improve proof generation and decomposition.
Our experiments hold the prover fixed to study how access to library declarations affects proof completion.

Library structure and prior attempts can also guide formal reasoning.
LEGO-Prover retrieves verified lemmas from a growing library~\citep{legoprover}, while RAutoformalizer and Aria use dependency information and retrieval during statement autoformalization~\citep{rautoformalizer,aria}.
For premise selection, \citet{graphpremise} incorporate dependencies into graph-aware representations.
\METHODNAME{} instead uses query-conditioned dependency traversal to expand the candidates considered by a semantic reranker.
During proof construction, COPRA records unsuccessful tactics and uses execution feedback to guide search~\citep{copra}; Reflexion more generally stores reflections on failed attempts~\citep{reflexion}.
Our optional failure memory associates unsuccessful repairs with recurring formal failure states, complementing the library evidence supplied by retrieval.

\subsection{Theorem-Proving Benchmarks}
\label{appd:related:proving_bench}

Formal proving benchmarks cover competition problems, textbook mathematics, and existing library developments.
miniF2F, FIMO, and PutnamBench evaluate competition-oriented mathematics~\citep{minif2f,fimo,putnambench}, while ProverBench adds a further collection of formalized problems, including AIME questions~\citep{deepseekprover2}.
ProofNet pairs undergraduate mathematical statements with formalizations~\citep{proofnet}, and FormalMATH broadens evaluation across mathematical domains and difficulty levels~\citep{formalmath}.
Domain-focused collections include CombiBench for combinatorics~\citep{combibench}, FATE for algebra at multiple difficulty levels~\citep{fate}, and MA-ProofBench for mathematical analysis~\citep{maproofbench}.

Other benchmarks emphasize the context and structure surrounding a theorem.
miniCTX evaluates proving within real Lean projects, where preceding definitions and lemmas supply relevant context~\citep{minictx}.
TheoremBench provides both main theorems and related supporting subtheorems to examine progress through a proof's structure~\citep{theorembench}.
Our proving evaluation uses FATE-H, MathlibMPR-Prop, and Lean-IMO-Bench to cover graduate algebra, Mathlib developments, and Olympiad problems~\citep{fate,leansearchv2,kung2026leap}.
Together with MathlibQR and MathlibMPR, these benchmarks connect declaration retrieval and premise coverage to proof completion under different reflection budgets.

\section{Implementation Details}
\label{appd:impl}

\subsection{Retrieval Backend and Dependency Graph}
\label{appd:impl:backend}

\paragraph{Corpus and graph representation.}
The implementation builds on the Mathlib~\texttt{v4.28.0-rc1} corpus released with LeanSearch~v2~\citep{leansearchv2}, retaining its declaration metadata, informal descriptions, and precomputed embeddings.
The released dependency graph contains 310,579 declarations and 4,528,330 forward edges extracted with \texttt{jixia}~\citep{jixia2024}.
The builder resolves structured references to declaration names, removes self-references and duplicate targets, and ignores references whose targets are absent from the graph.
Its dependency-only profile uses forward dependency edges with unit weights.
The transition matrix is stored in compressed sparse row format, with a separate indicator for dangling nodes.
Inclusion of theorem proof-body references is controlled at graph-construction time and recorded in the graph metadata.

\paragraph{Semantic retrieval and relevance scores.}
Queries are encoded with Qwen3-Embedding-8B~\citep{qwen3embedding} and $\ell_2$-normalized before cuVS CAGRA search.
Qwen3-Reranker-8B scores each query--declaration pair using the final-position logits for the tokens \texttt{yes} and \texttt{no}:
\begin{equation}
    q_i = \frac{\exp(z_{i,\mathrm{yes}})}
    {\exp(z_{i,\mathrm{yes}})+\exp(z_{i,\mathrm{no}})}.
    \label{eq:impl:relevance}
\end{equation}
These nonnegative scores define the seed weights in \cref{sec:method:retrieval}.
Both reranking passes use the same checkpoint and query, with each formatted pair limited to 8,192 tokens.

\paragraph{Matched retrieval budgets.}
In all mathematical retrieval and theorem-proving evaluations, both retrievers return at most $K=50$ declarations per query, and \METHODNAME{} uses $N=200$ PPR-ranked expansion candidates.
For \METHODNAME{}, $K$ specifies both the number of semantic seeds and the number of declarations returned after final ranking, while $N$ specifies the number of PPR-ranked graph candidates considered for expansion.
LeanSearch~v2 also returns its top 50 declarations for each query.
Thus, the paired comparisons use the top-50 results of each retriever; graph expansion increases the internal candidate pool without increasing the per-query output budget.
The same setting applies to direct benchmark queries, reasoning-based subqueries, and reflection-time queries.
Multi-query aggregation is a separate operation and can combine declarations from several top-50 lists.

\paragraph{Propagation and candidate refinement.}
The retrieval configuration uses 50 semantic seeds, 200 PPR-ranked graph candidates, and a refined candidate pool of at most 100 declarations.
These internal budgets are separate from the per-query output budget of $K=50$.
PPR uses restart probability $r=0.15$, at most 50 iterations, and an $\ell_1$ convergence tolerance of $10^{-6}$.
The computation uses double precision and returns dangling mass to the seed distribution.

Before the second reranking pass, the implementation merges the first-pass semantic candidates with the PPR-ranked candidates and applies a graph-aware prefilter:
\begin{equation}
    h_i = \alpha\,\operatorname{mm}(q_i)
          +\beta\,\operatorname{mm}(p_i)
          -\gamma\log\!\left(1+\deg_{\mathrm{in}}(i)
                                +\deg_{\mathrm{out}}(i)\right),
    \label{eq:impl:prefilter}
\end{equation}
where $\operatorname{mm}$ denotes min--max normalization over the candidate union.
New graph-only candidates receive an initial semantic score of zero.
The shared prefilter defaults are $\alpha=\beta=1$ and $\gamma=0.05$.
The retained candidates then receive new semantic scores from the reranker.
The \texttt{qwen\_\allowbreak{}ppr\_\allowbreak{}zscore} option implements the final score in \cref{sec:method:retrieval} with $\lambda=0.005$; its mean and standard deviation are computed over the retained candidates for the current query.
A zero-variance PPR vector contributes a zero correction.

\subsection{Search Loop Policy}
\label{appd:impl:search_policy}

\paragraph{Single-query retrieval.}
LeanSearch~v2 and \METHODNAME{} share the same query and result interface.
Each result contains a declaration name, module, kind, formal signature, and available informal description.
The client explicitly specifies whether graph augmentation is enabled, allowing both retrieval conditions to use the same serving backend.
The proving and reasoning callers retain at most $K=50$ declarations from each query, even when graph retrieval uses a larger internal pool.

\paragraph{Reasoning-based retrieval.}
The reasoning controller follows a decompose--retrieve--filter--judge loop~\citep{leansearchv2}.
Given the formal statement and an available informal description, the decomposition model generates queries for definitions, high-level results, and individual proof steps.
Duplicate query strings are removed before retrieval.
A filtering model selects relevant declarations for each subquery using the target statement and the corresponding step description.
A judging model then evaluates whether the resulting premises support the proposed plan.
An accepted plan terminates the loop; otherwise, the plan and judge feedback are supplied to the next decomposition call.
MathlibMPR retrieval and reasoning-based pre-retrieval for theorem proving use the same four templates, reproduced in \cref{appd:prompt:reasoning}.

After filtering, declarations are identified by their module and fully qualified name and ranked by the sum of their positional discounts:
\begin{equation}
    a(d)=\sum_{u\in\mathcal{Q}:\,d\in F_u}
         \frac{1}{\log_2(\operatorname{rank}_{F_u}(d)+1)},
    \label{eq:impl:aggregation}
\end{equation}
where $F_u$ is the filtered list for subquery $u$ and ranks start at one.
The highest-scoring declarations form the final reasoning context.
The multi-branch controller runs independent plans concurrently and uses the first branch accepted by the judge.
If no branch is accepted, it pools the final filtered lists from the completed branches and applies the same aggregation rule.

\paragraph{Integration with proof refinement.}
In Standard Mode, the first proof attempt receives no retrieved context.
After a failed attempt, a query generator uses the target statement, current candidate, and Lean diagnostic to request supporting declarations.
The reported proving setup requests five queries per retrieval action and retains up to 50 declarations per query.
The five lists contain at most 250 entries and are deduplicated in first-seen order before being supplied to the prover.
Retrieved entries are formatted as Lean declarations with their informal descriptions.

Reasoning-based proving first obtains an initial context through the reasoning controller.
The reported Reasoning Mode uses the adaptive policy: the initial proof sketch and retrieved premises are supplied to the prover, and subsequent failures can trigger the same reflection-time queries used in Standard Mode.
The initial premises are combined with newly retrieved declarations after failures.
The fixed-context \texttt{reasoning}/\texttt{graph\_\allowbreak{}reasoning} paths also exist in the codebase, but the reported adaptive policy corresponds to their \texttt{\_adaptive} variants.
The \texttt{GET\_\allowbreak{}QUERY\_\allowbreak{}INIT} template is not invoked in the five evaluated system configurations.

\subsection{Mathematical Retrieval Setting}
\label{appd:impl:retrieval}

\paragraph{MathlibQR.}
The evaluation expands each benchmark declaration into its available query variants and selects the shared-declaration subset using the released list of 171 names.
This yields the 810 queries used in \cref{sec:exp:retrieval}.
Dense retrieval supplies 200 candidates for the initial reranking stage; the matched comparison uses the final top 200 declarations from each retriever, with $K=50$ and $N=200$ for \METHODNAME{}. Compared to \cref{fig:method}, the last step of choosing final top-$K$ is not used, keeping the full list for deeper comparison.
With one annotated target per query, Recall@$k$ is the indicator that the target appears within the first $k$ results, and nDCG@$k$ is its logarithmically discounted rank when retrieved.
Metrics are averaged across queries.

\paragraph{MathlibMPR.}
The reasoning input consists of the formal statement and its \texttt{NL\_\allowbreak{}main\_\allowbreak{}result} description; the supplied informal-proof field is empty.
The released driver defaults to two reasoning branches, with one initial planning pass and up to three revisions per branch.
Each subquery retrieves up to 50 declarations for relevance filtering under the shared $K=50$, $N=200$ setting.
The judge examines at most 30 unique filtered declarations, and the final aggregated ranking contains at most 100 declarations.
For failed filter parsing, the implementation retains the first five candidates; failed judge parsing triggers another planning iteration when budget remains.

Evaluation normalizes retrieval identifiers to declaration names and checks whether each annotated premise group is hit by at least one retrieved member.
Group recall is macro-averaged across problems.
Covered@$k$ checks complete coverage of the original route or the annotated alternative route.
Both metrics are evaluated at $k\in\{5,10,20,30,50\}$.
As described in \cref{sec:exp:retrieval_setting}, the reported LeanSearch~v2 and \METHODNAME{} results are averaged over Gemini~3.1~Pro and Claude~Sonnet~5.

\subsection{Theorem Proving Setting}
\label{appd:impl:proving}

\paragraph{Problems and model calls.}
The proving evaluation uses 100 FATE-H problems, the 50 MathlibMPR entries selected by the released proposition-ID list, and 30 problems from each Lean-IMO-Bench difficulty level~\citep{fate,leansearchv2,kung2026leap}.
Benchmark loaders retain the formal statements and required source context.
The Olympiad loader supplies the problem description and Lean statement while keeping solution and grading fields outside the model input.
Gemini~3.1~Pro is the prover backbone in the reported experiments.
The implementation requests temperature zero for proof generation, query generation, and the reasoning-controller calls.
Each revision is conditioned on the current candidate, latest diagnostic, and available retrieval and failure-memory context, maintaining a single proof trajectory.

\paragraph{Verification and proof-hole integrity.}
The Lean verifier uses \texttt{lean-interact} with a Mathlib environment, with \texttt{v4.28.0-rc1} as the fallback toolchain version.
Its default timeout is 600 seconds per verification call.
The integrity-aware path isolates one target proof hole and preserves the surrounding source as an immutable prefix and suffix.
The model returns a tactic body, which the host inserts into this source.
Before compilation, the checker rejects changes to the protected context, additional top-level commands, and prohibited constructs such as \texttt{sorry}, \texttt{admit}, and newly introduced \texttt{axiom} declarations.
The checks ignore comments and string contents.
Under this contract, success requires both valid source boundaries and successful Lean elaboration without unresolved \texttt{sorry} obligations.

\paragraph{Reflection budgets.}
The implementation records the initial attempt at round zero and subsequent revisions with increasing round indices.
The reported evaluation measures cumulative completion at budgets $k\in\{8,16,32\}$.
Proof-generation calls, retrieval actions, and compiler submissions are tracked separately in the budgeted controller.
One retrieval action can issue several search queries, while initial reasoning preparation has its own planning budget.
These counters also distinguish a newly compiled candidate from an exact failed candidate rejected using cached feedback.

\paragraph{Failure memory.}
Memory is initialized separately for each proving problem and is not used in the standalone mathematical retrieval benchmarks.
For a stable failed candidate, the implementation serializes its error category, normalized diagnostic, unresolved goals, local context, inferred failing tactic, referenced identifiers, and nearby source snippet.
A SHA-256 hash identifies the state, and the record stores its occurrence count and associated unsuccessful edits~\citep{sha-256}.
The repeat threshold is two observations.
Feedback includes at most three recurrent states and the three most recent failed edits for each, together with a directive to change the unsuccessful repair.

When memory is enabled, the \texttt{FAILURE\_\allowbreak{}MEMORY\_\allowbreak{}GUIDANCE\_\allowbreak{}STRICT} instruction is included in the initial proof prompt and every proof-refinement prompt.
The structured \texttt{failure\_\allowbreak{}memory} summary is appended to compiler-tool feedback, so both the proof-refinement prompt and the reflection-query prompt receive it.
The guidance treats a recorded tactic or declaration as evidence about a failed application in its context, rather than a global prohibition on reusing that tool.
A corrected application may remain valid after changing its arguments, goal, instances, or local hypotheses.
When a state occurs at least twice, the summary includes the \texttt{DEAD\_\allowbreak{}END\_\allowbreak{}REPEATED} directive.
The complete guidance and repeated-state directive are reproduced in \cref{appd:prompt:memory}.

Exact candidates with cached deterministic failures are rejected without another compiler call; their generation calls still count toward model usage.
The already prepared retrieval context is retained for the replacement candidate.
Repeated search queries are cached by a formal-state fingerprint and normalized query string, and cached results are reused.
Infrastructure failures are excluded from the stable-failure cache.
These operations add feedback and eliminate duplicate work while preserving the single-trajectory proving loop.

\paragraph{API cost accounting.}
API expenditure includes paid calls for proof generation, query generation, and reasoning-based retrieval, including calls made on unsuccessful problems.
Relative costs are normalized by the no-retrieval configuration within the same benchmark and memory setting.
For the aggregate comparison, cost per solved problem is
\begin{equation}
    \operatorname{CostPerSolved}
    =\frac{\sum_{i=1}^{210}C_i}
           {\sum_{i=1}^{210}\mathbbm{1}
           \{i\text{ is solved at budget }32\}},
    \label{eq:impl:cost}
\end{equation}
where $C_i$ is the API expenditure on problem $i$.
This measure covers API usage; local retrieval and Lean-verification hardware costs are outside its definition.
The result records retain proof attempts, compiler diagnostics, retrieval traces, and available usage and memory statistics for inspecting individual trajectories.

\section{Case Study}
\label{appd:case}

\subsection{Mathematical Retrieval}

\paragraph{Dependency references improve retrieval quality.}
On MathlibQR, semantic search often identifies an object's constructor before identifying the object itself. For the query describing ``a family of morphisms between two functors that commutes with morphism action,'' LeanSearch~v2 retrieves the constructor \texttt{NatTrans.mk} but does not place \texttt{CategoryTheory.\allowbreak{}NatTrans} in its top 100. A type-reference edge from the constructor introduces the parent declaration, which \METHODNAME{} ranks third. The same pattern appears for a query about bases and independent sets satisfying the exchange axioms: \texttt{Matroid.mk} leads to \texttt{Matroid}, ranked second by \METHODNAME{} but absent from LeanSearch~v2's top 100. For a description of a category equipped for Quillen-style homotopy theory, \texttt{ModelCategory.\allowbreak{}mk'} similarly leads to \texttt{HomotopicalAlgebra.\allowbreak{}ModelCategory}, which \METHODNAME{} ranks first. In each case, the semantically recognizable API entry provides a path to the underlying mathematical object.

References in declaration bodies also expose a complementary route. For the MathlibQR query expressing a presheafed space as $(X,\mathcal{F}:\mathrm{Open}(X)^{\mathrm{op}}\to C)$, a semantically retrieved global-sections restriction functor refers to \texttt{AlgebraicGeometry.\allowbreak{}PresheafedSpace} in its value. Graph expansion recovers the latter at rank four, whereas LeanSearch~v2 does not return it in the top 100. For the query defining a scheme as a locally ringed space locally isomorphic to an affine spectrum, a value reference from \texttt{AlgebraicGeometry.\allowbreak{}AffineScheme} connects to \texttt{AlgebraicGeometry.\allowbreak{}Scheme}; the target moves from outside the baseline top 100 to rank eight. Finally, for ``morphism between functors,'' the implementation of \texttt{CategoryTheory.\allowbreak{}Functor.\allowbreak{}category} refers to natural transformations. Following that value reference recovers \texttt{CategoryTheory.\allowbreak{}NatTrans} at rank 14, again from outside the baseline top 100. These examples show how implementation-level relationships can bring a relevant declaration into the candidate set when its own description is a weak semantic match.

\paragraph{Better retrieval improves premise-route coverage.}
The benefit on MathlibMPR is visible in the composition and ordering of the filtered premise list. In MathlibMPR Query 15, the angle argument requires both \texttt{SSameSide.\allowbreak{}oangle\_\allowbreak{}sign\_\allowbreak{}eq} and \texttt{angle\_\allowbreak{}eq\_\allowbreak{}abs\_\allowbreak{}oangle\_\allowbreak{}toReal}. LeanSearch~v2 places them at ranks one and seven; \METHODNAME{} places them at ranks one and four, completing the two-premise route within the top five. In MathlibMPR Query 75, a route through cover lifting, Grothendieck-topology closure, and adjunction naturality is similarly more concentrated. \METHODNAME{} places \texttt{IsCocontinuous.\allowbreak{}cover\_\allowbreak{}lift} at rank two rather than 18 and \texttt{superset\_\allowbreak{}covering} at rank three rather than eight. Its filtered context also contains both the left- and right-naturality lemmas for the adjunction's hom equivalence, at ranks 21 and 15; neither appears in the corresponding LeanSearch~v2 filtered list. The resulting context brings the principal steps of this diagrammatic route together.

\subsection{Theorem Proving}

\paragraph{\METHODNAME{} improves proof success rate within budget.}
For MathlibMPR-Prop Problem 51, LeanSearch~v2 exhausts the budget, while \METHODNAME{} succeeds at reflection round 10. The successful trajectory searches for Radon--Nikodym chain rules, density transformations, and composition products, then combines the corresponding measure-theoretic lemmas in its accepted proof. On FATE-H Problem 88, the LeanSearch~v2 trajectory again ends at round 31; the \METHODNAME{} trajectory succeeds at round 10 after developing the route from regularity in a quotient to ideal membership. On FATE-H Problem 100, \METHODNAME{} succeeds at round 18 using a square-zero extension and the nilpotence of its module component, whereas LeanSearch~v2 exhausts round 31. These cases illustrate how a search-guided trajectory can reach a compiler-accepted construction within the available budget.

\paragraph{\METHODNAME{} also helps complete proofs in fewer reflection rounds.}
Both systems solve MathlibMPR-Prop Problem 34, but \METHODNAME{} succeeds at round seven rather than 16. Its proof develops the colon-ideal argument using, among other facts, \texttt{Submodule.\allowbreak{}mem\_\allowbreak{}colon\_\allowbreak{}singleton}; total API cost falls from \$8.33 to \$3.25. For MathlibMPR-Prop Problem 66, the route through multiplicative-height bounds and finite products reaches a proof at round five rather than 16, reducing cost from \$9.37 to \$2.85. On FATE-H Problem 89, queries about flatness, going down, and prime-ideal height align with the accepted proof's height argument. \METHODNAME{} completes it at round seven rather than 21, at \$3.24 rather than \$10.82.

\paragraph{Graph retrieval lowers API cost per attempt.}
Fewer attempts account for part of the savings, but the observed API cost per attempt also decreases. For instance, on MathlibMPR-Prop Problem 34, the cost decreases from \$0.490 to \$0.406 per attempt. On FATE-H Problem 15, a proof that $\mathbb{Z}[\sqrt{-5}]$ is not a principal ideal ring, decreases from \$0.572 to \$0.415 per attempt while finishing at round three rather than seven. On Lean-IMO-Bench Advanced Set Problem 25, a digit-bound argument, the corresponding figures are \$0.543 and \$0.416, with completion at round three rather than six. Thus the shorter trajectories in these cases also require less billed model usage per attempt.

\paragraph{Failure memory improves proving efficiency.}
Failure memory supplies an additional source of improvement. On Lean-IMO-Bench Basic Set Problem 16, the no-memory \METHODNAME{} run exhausts round 31. The memory-enabled run records a repeated recursion-depth failure, changes its subsequent proof action, and closes the remaining arithmetic contradiction with \texttt{omega} at round nine. On FATE-H Problem 12, both runs succeed, but enabling memory reduces completion from round 30 to round seven and API cost from \$17.26 to \$4.56. The trace records the intervening compiler errors, including type mismatches, before the accepted revision.

\section{LLM Prompting and Sampling}
\label{appd:prompt_sampling}

\subsection{Prompt Interfaces}
\label{appd:prompt:interfaces}

We use the same prompt templates in matched LeanSearch~v2~\citep{leansearchv2} and \METHODNAME{} comparisons.
Retrieved declarations are presented through an informal description, when available, followed by the formal declaration or signature.
Our prompt templates are modified from the templates of \citet{leansearchv2}, correcting some parts that may lead to incomplete proofs. See \cref{appd:prompt:reasoning,appd:prompt:proving} for our prompts.

\paragraph{Mathematical retrieval.}
MathlibQR uses the benchmark queries directly and requires no LLM-generated queries.
For MathlibMPR, the reasoning model receives the formal theorem and its informal description.
The reference-proof field is left empty, and gold premise groups are used only for evaluation.
The same reasoning prompts are used to prepare the initial context for theorem proving in Reasoning Mode.

Reasoning retrieval follows four stages: planning, retrieval, filtering, and route assessment.
The planning prompt first seeks high-level theorems that may directly advance the proof, then key definitions and lemmas for individual steps, with at most 20 queries across all categories.
For each query, a filtering prompt examines up to 50 declarations and selects results according to their mathematical relevance, Lean types, and intended role in the proof.
The filter may return an empty list.
A judge assesses whether the filtered premises support an implementable Lean proof route and returns a \texttt{good} or \texttt{bad} judgment with feedback.
This judgment evaluates the proposed route; successful theorem proving still requires Lean verification.

Two reasoning branches run in parallel, each allowing an initial plan and up to three revisions based on judge feedback.
The first branch judged \texttt{good} supplies the final route.
If neither branch is accepted, their final filtered lists are pooled and declarations are ranked by the sum of their position-discounted contributions across queries.
The resulting list contains at most 100 declarations.

\paragraph{Theorem proving.}
Standard Mode begins with a proof attempt without retrieved context.
Reasoning Mode first obtains a proof sketch and filtered premises through the reasoning procedure above.
After a failed attempt, both modes permit reflection-time retrieval: the query prompt receives the theorem, the preceding tactic body, and the compiler feedback, and requests exactly five complementary natural-language queries.
The next proof attempt is conditioned on the previous attempt, compiler feedback, and deduplicated retrieval results.
There is no separate initial-query generation step outside the reasoning procedure.

The proof-generation prompts supply an immutable theorem prefix and suffix and request a complete replacement tactic body.
Shared proof-integrity instructions require the prover to establish the target from its hypotheses and existing library declarations, with any auxiliary results proved locally.
The generated body is inserted into the fixed source context and checked for source integrity before Lean verification.
Source-integrity errors and Lean diagnostics are returned as feedback for the next revision.

\paragraph{Failure memory.}
Failure memory is used only in theorem proving.
When enabled, its guidance is included in the initial proof prompt and every revision prompt, while the memory itself is initially empty.
After a failed attempt, a structured failure summary is appended to the compiler feedback and is therefore available to both proof revision and reflection-time query generation.
The summary records recurring formal states, occurrence counts, and unsuccessful tactic or declaration applications.
An unchanged candidate with a cached deterministic failure is blocked without another compilation.

\subsection{Decoding and Search Budgets}
\label{appd:prompt:budgets}

\paragraph{Model settings.}
Gemini~3.1~Pro with \textsc{High} thinking is used for all proving runs and one MathlibMPR retrieval run.
Requests set \texttt{top\_p} to 1 and cap model output at 65,536 tokens.
Temperature is left at the provider default, and Google Search is disabled.
The second MathlibMPR retrieval run uses Claude~Sonnet~5 with adaptive thinking, high effort, and a 128,000-token output cap.
Its temperature and \texttt{top\_p} remain at the provider defaults, and no sampling seed is supplied.
Each paired comparison uses the same model settings and prompt templates.

\paragraph{Retrieval budgets.}
All retrieval calls in the retrieval and proving evaluations use $K=50$ returned declarations, and \METHODNAME{} uses $N=200$ PPR-ranked expansion candidates.
Thus, the comparison uses the top 50 results from each retriever under the same per-query output budget.
In MathlibMPR retrieval and initial reasoning retrieval for proving, each query supplies up to 50 declarations to the filter; the filtered lists are aggregated into at most 100 declarations.
Each reflection-time retrieval action generates five queries with up to 50 results each, which are deduplicated before being passed to the prover.
The larger internal candidate pool used by graph retrieval does not increase the per-query output budget.

\paragraph{Proving budgets.}
A trajectory contains one initial candidate and at most 31 compiler-tested revisions.
At most one retrieval action is available before each revision, for a maximum of 31 actions.
Query generation and proof generation share a 63-call proof-stage model budget.
Initial reasoning retrieval has a separate budget and does not reduce the available proof-stage calls.
Each problem follows a single candidate trajectory rather than selecting among multiple parallel proofs.

\subsection{Instance Selection and Accounting}
\label{appd:prompt:accounting}

The retrieval evaluations use the fixed 810-query fair subset of MathlibQR and all 69 MathlibMPR instances.
The proving evaluations use identical problem sets across configurations: 50 MathlibMPR-Prop propositions, 100 FATE-H problems, and 30 problems from each of the Basic and Advanced Lean-IMO-Bench sets.
Each configuration is evaluated with one proving trajectory per problem.

We report pass@$k$ as cumulative success along that trajectory by reflection round $k$.
The initial compiler submission is indexed by zero, and retrieval calls do not advance the round counter.
Under the 32-submission cap, pass@$32$ is the full-budget result.
API expenditure is calculated from response-reported token usage for query generation, initial reasoning retrieval, filtering, judging, and proof generation, including calls on unsuccessful problems.
Local embedding, reranking, graph traversal, and Lean compilation are excluded from this API-cost measure.

\lstdefinestyle{LGAppendixPrompt}{
  language={},
  basicstyle=\ttfamily\fontsize{8}{9.4}\fontencoding{T1}\selectfont,
  columns=fullflexible,
  keepspaces=true,
  breaklines=true,
  breakatwhitespace=false,
  breakautoindent=true,
  breakindent=1em,
  showstringspaces=false,
  showspaces=false,
  showtabs=false,
  upquote=true,
  numbers=none,
  frame=single,
  framerule=0.3pt,
  framesep=5pt,
  xleftmargin=5pt,
  xrightmargin=5pt,
  aboveskip=0.65\baselineskip,
  belowskip=0.75\baselineskip,
  tabsize=2,
  literate={¬}{{\ensuremath{\neg}}}1
           {ℤ}{{\ensuremath{\mathbb{Z}}}}1
           {√}{{\ensuremath{\surd}}}1
           {²}{{\ensuremath{{}^{2}}}}1
           {−}{{\ensuremath{-}}}1
           {σ}{{\ensuremath{\sigma}}}1
}

\subsection{Reasoning Retrieval Prompts}
\label{appd:prompt:reasoning}

The following prompts are shared by MathlibMPR retrieval and initial reasoning retrieval for theorem proving.
Fields in braces denote instance-specific inputs.
\Needspace{9\baselineskip}
\subsubsection{Initial route generation.}
\label{appd:prompt:decompose_initial}
\begin{lstlisting}[style=LGAppendixPrompt]
You are a mathematician that is very familiar with abstract algebra and commutative algebra, and their corresponding Lean 4 formalizations and Mathlib infrastructure.
Your task is to generate several queries to find key Lean 4 items for the formal proof. **Crucially, you should first attempt to find "High-Level Theorems" that might directly imply the conclusion or cover a significant portion of the informal proof.** Mathlib often contains powerful generalizations (e.g., "PID implies UFD") that can solve specific problems in 1-2 steps, avoiding the need to reconstruct the proof from scratch.
Therefore, your queries should target three categories:
1. **High-Level Targets:** Queries for theorems that might bridge the gap between assumptions and conclusion directly.
2. **Key Definitions:** The structures described in the statement (to ensure the downstream LLM understands their exact meaning and usage).
3. **Step-wise Lemmas:** The specific tools needed for the detailed proof steps (to avoid proving them from scratch).
These queries will be fed into LeanSearch, a semantic search engine for Mathlib4, to find the corresponding Lean 4 items in Mathlib4. The following guidelines for the query generation task will help you to enhance the search quality.

**Query Generation Guidelines:**
1. The queries should be written in natural language, and contain enough details about the item you are searching for.
   a) For definitions/structures, you should clearly state that you are searching for a definition in the query ("The definition of ..."), and state the name (if it exists) and the full content of this definition.
   b) For lemmas/theorems/instances, you should state the name (if it exists) and the full content of this lemma/theorem/instance.
2. Note that you are searching for items in Mathlib4, a formal language library. It doesn't contain all possible definitions/theorems. It is built in a Bourbaki-style, and only contains the most important and frequently used definitions/theorems.
   a) Do not expect a theorem for a specific choice of a constant exists in Mathlib4. For example, when dealing with `theorem not_isPrincipalIdealRing_Zsqrtd_neg_five : ¬IsPrincipalIdealRing (Zsqrtd (-5))`, "How to use the norm to analyze divisibility in Zsqrtd (-5)" is not a good query. Instead, you should search for "The definition of the norm on `ℤ√d`. For an element `n = x + y√d` in `ℤ√d`, the norm is defined as `n.norm = n.re * n.re - d * n.im * n.im`, i.e. `x² − d y²`".
   b) For problems that contain specific calculations (e.g., structure of Zsqrtd (-5), group of order 336 is not simple group), usually there is no direct applicable theorem in Mathlib4. Instead, you need to search for tools that can help in the calculations.
   
## Theorem to Prove

{formal_statement}

## Informal Statement

{informal_statement}

## Informal Proof (Reference)

{informal_proof}

## Your Task

Generate a complete proof plan with queries organized into three categories: definitions, high-level theorems, and step-by-step proof.

**IMPORTANT: You should carefully analyze the informal proof provided above and use it as the foundation for your step-by-step decomposition.** Break down the informal proof into clear, logical steps that can be directly translated into Lean 4 code. Each step should capture a single inference from the informal proof.
**Constructive Concreteness:** If a step claims an object exists (e.g., an inverse, a bounding constant, a specific element), explicitly state how to construct or obtain it. Avoid vague appeals to named techniques without specifying the concrete algebraic operations involved.
**Typeclass & Coercion Awareness:** When the proof involves multiple related structures (e.g., `Ideal` vs `FractionalIdeal`, `Polynomial R` vs `MvPolynomial σ R`), explicitly identify the type boundaries and plan bridge lemmas for transitions between them.
**You must still provide detailed step-by-step proof as a fallback**, in case the high-level theorems are not found. The steps should be a coherent proof following the structure of the informal proof, and for each step, determine if it is:
- **Type A) Easily provable with a theorem:** Propose 1-3 queries for the lemma/theorem/instance. Usually, one query is enough if your step is clear and simple.
- **Type B) Requires calculation:** The query number can be more than 3. Try to find effective calculation tools.
Try to rearrange the proof steps so that you can use powerful theorems in Mathlib4 to accelerate the proof.

## Response Format

Provide your response in the following JSON format:
```json
{
  "definition_queries": [
    "The definition of a Unique Factorization Monoid (UFD), also known as UniqueFactorizationMonoid.",
    "The definition of the height of a prime ideal, Ideal.height."
  ],
  "highlevel_queries": [
    "Theorem stating that a Noetherian domain where every height 1 prime is principal is a UFD.",
    "Nagata's criterion for UFDs involving prime elements or height 1 primes."
  ],
  "steps": [
    {
      "description": "Show that a Noetherian integral domain is a UFD if and only if every irreducible element is prime",
      "reasoning": "This simplifies the main goal, as the existence of factorization is guaranteed by IsNoetherianRing",
      "queries": ["Theorem stating that a Noetherian integral domain is a UFD if and only if every irreducible element is prime."]
    },
    {
      "description": "Apply Krull's Principal Ideal Theorem to the minimal prime over (x)",
      "reasoning": "For an irreducible element x, we need to find a prime ideal P minimal over (x) with height 1",
      "queries": ["Krull's Principal Ideal Theorem: In a Noetherian ring, any prime ideal minimal over a non-zero principal ideal has height 1."]
    },
    {
      "description": "...",
      "reasoning": "...",
      "queries": ["..."]
    }
  ]
}

```
**Important notes:**
- `definition_queries`: 1-3 queries for core definitions (optional, can be empty array)
- `highlevel_queries`: 2-4 queries for high-level theorems that might solve the problem directly (optional, can be empty array)
- `steps`: 3-7 major proof steps, each with description, reasoning, and queries
- Total number of queries should be no more than 20
- By experience, if the theorem doesn't need calculation, usually 3-7 queries is enough

Now generate the proof plan:
\end{lstlisting}
\Needspace{9\baselineskip}
\subsubsection{Route revision after judge feedback.}
\label{appd:prompt:decompose_reflect}
\begin{lstlisting}[style=LGAppendixPrompt]
You are a mathematician that is very familiar with abstract algebra and commutative algebra, and their corresponding Lean 4 formalizations and Mathlib infrastructure.

Your previous proof plan led to a formal sketch that was rejected as inadequate. You need to rethink the overall proof strategy and generate a COMPLETELY NEW proof plan.

## Theorem to Prove

{formal_statement}

## Informal Statement

{informal_statement}

## Informal Proof (Reference)

{informal_proof}

## Previous Proof Plan (REJECTED)

{previous_steps}

## Quality Judge's Feedback

{quality_reasoning}

## Your Task

Generate a COMPLETELY NEW proof plan with a different approach. Do NOT just tweak the previous plan - rethink the entire strategy.

**IMPORTANT: Re-read the informal proof carefully and consider a fundamentally different way to decompose it into formal steps.** The previous decomposition was rejected, so try a different structural approach while still following the overall narrative of the informal proof.

**Query Generation Guidelines:**
1. The queries should be written in natural language, and contain enough details about the item you are searching for.
   a) For definitions/structures, clearly state you are searching for a definition ("The definition of ...").
   b) For lemmas/theorems/instances, state the name (if it exists) and the full content.
2. Remember that Mathlib4 is built in a Bourbaki-style and only contains the most important and frequently used definitions/theorems.
   a) Do not expect theorems for specific constants (e.g., Zsqrtd (-5)). Instead, search for general tools and definitions.
   b) For problems with specific calculations, search for calculation tools rather than direct theorems.
**Strategy:**
- **Try a fundamentally different proof structure** - if previous plan focused on direct calculation, try finding high-level theorems; if it relied on general theorems, try a more constructive approach
- **Different decomposition:** If previous plan had N steps, consider N+1 or N-1 steps
- **Look for alternative mathematical pathways** that might be better supported by Mathlib
- **Prioritize finding high-level theorems** that can bridge large portions of the proof

## Response Format

Provide your response in the following JSON format:

```json
{
  "definition_queries": [
    "The definition of ..."
  ],
  "highlevel_queries": [
    "Theorem stating that ..."
  ],
  "steps": [
    {
      "description": "...",
      "reasoning": "...",
      "queries": ["..."]
    }
  ]
}
```

**Important notes:**
- `definition_queries`: Core definitions (can be empty if they were already found)
- `highlevel_queries`: 2-4 queries for completely different high-level approaches
- `steps`: New step-by-step proof with fundamentally different structure
- Total queries should be no more than 20
- Focus on finding theorems that can be easily verified with Lean

Now generate the NEW proof plan:
\end{lstlisting}
\Needspace{9\baselineskip}
\subsubsection{Per-query relevance filtering.}
\label{appd:prompt:filter}
\begin{lstlisting}[style=LGAppendixPrompt]
You are an expert in formal mathematics and Lean 4. Your task is to filter search results for a specific query, keeping only the most relevant and useful items.

## Context

We are proving the following theorem:
{formal_statement}

## Query Context

This query was generated for proof step: "{step_description}"
Query: "{query}"

## Search Results

{search_results}

## Important Notes

The retrieval process is carried out by an embedding and reranking based search engine. It is excellent at finding semantically relevant documents, but **unable to handle cases where the document in Mathlib4 is not in its common form in natural language** (e.g., using module theory to describe the minimal polynomial of a matrix, or the difference between `Fintype` and `Finite`).

## Your Task

Further filter the retrieved results by selecting documents that are **truly relevant** to the search query.
**A document is relevant if:**
1. It matches the mathematical concept described in the query
2. Its **Type Signature** matches the expected mathematical objects
   - Pay special attention to type differences (e.g., `Nat` vs `Int`, `Fintype` vs `Finite`, `Ideal R` vs `Submodule R R`)
   - Check if the theorem applies to the specific algebraic structures in our target theorem
3. It could directly help prove the step (not just tangentially related)
**A document should be FILTERED OUT if:**
1. It is about a different mathematical concept (even if the name sounds similar)
2. Its type signature does not match our needs (wrong types, wrong structure)
3. It is an internal implementation detail or auxiliary lemma
4. It is redundant with other better results
**Important:** If there is no direct match in the search results, especially for theorem queries (as opposed to definition queries), **you should output an empty list `[]`**. It is better to have no results than to keep marginally relevant items that will confuse the downstream proof generation.
## Response Format

Provide your response in JSON format:

```json
{
  "kept_results": [0, 2],
  "reasoning": "Result 0 provides the exact theorem for Noetherian domains with the correct type signature (CommRing R, IsDomain R). Result 2 gives the definition of height with the expected type Ideal.height. Result 5 is filtered out because it applies to Submodule rather than Ideal, which is a type mismatch."
}
```
Where:
- `kept_results`: A list of indices (0-based) of results to keep. **Can be empty `[]` if no results match.**
- `reasoning`: Explain your filtering decisions, especially mentioning **which results are kept and why**, as well as **which results are filtered out and why** (focusing on type signature mismatches).

Now evaluate the results:
\end{lstlisting}
\Needspace{9\baselineskip}
\subsubsection{Route-quality assessment.}
\label{appd:prompt:judge}
\begin{lstlisting}[style=LGAppendixPrompt]
You are an expert in Lean 4 theorem proving. Judge whether the current proof plan is suitable for decomposition into Lean subgoals that remain tractable in the downstream proof-completion stage (after sketch generation).

## Theorem to Prove

{formal_statement}

## Informal Proof Plan

{informal_plan}

## Available Theorems (filtered from search)

{available_theorems}

## Filtering Reasoning (optional)

{filter_reasoning}

## Judging Criteria

You are judging **pre-sketch plan quality for downstream proof completion**, not final proof correctness.

**CRITICAL**: This is a task of **writing Lean code**, not just mathematics. Judge difficulty by **Lean implementation complexity**, not mathematical elegance.
Evaluate the informal plan **step by step** (not as a single overall impression):
- For each step, check whether the step has a clear, verifiable conclusion that can map to a Lean subgoal (`have ... : ...`).
- For each step, check whether it is mathematically sound (not false, not self-contradictory, not based on invalid implications).
- For each step, estimate whether it is likely lightweight or heavy in Lean implementation.
- For each step, explicitly judge Mathlib support **based on the retrieved-and-filtered theorems listed above** (`Available Theorems`).
- Do not rely on assumed library knowledge that is not reflected in the retrieved-and-filtered results; if support is missing in those results, treat it as a coverage gap.
- Consider: Can this step be resolved with available theorems + simple tactics (`rfl`, `simp`, `ring`)?
- If a step is likely heavy, check whether it should be split into smaller bridge steps.
Typical Lean-hard patterns to watch for (even when math is simple):
- hidden coercion/casting jumps;
- bundled "expand + coefficient compare + contradiction" in one step;
- omitted bridge lemmas between algebraic transformations;
- assumptions that are mathematically plausible but awkward in Lean without extra setup.
A plan is **GOOD** if:
1. Most key steps have at least one relevant theorem/definition support,
2. Retrieved theorems have acceptable type/structure alignment with target steps,
3. There is no obvious coverage hole for critical bridge steps,
4. The overall decomposition seems implementable in Lean sketch form,
5. No single step appears disproportionately difficult without decomposition.
6. No mathematically incorrect key step is present.
A plan is **BAD** if:
1. Critical steps have no meaningful support,
2. Retrieved results are mostly tangential or mismatched,
3. The plan likely needs query/strategy revision before sketching,
4. One or more steps are too coarse (especially if they merge multiple hard transformations without intermediate conclusions).
5. Any mathematically incorrect key step is present.

**Be REALISTIC but not overly strict.** If most steps look achievable but one is borderline, consider the overall balance.

## Response Format

```json
{
  "judgment": "good" or "bad",
  "reasoning": "Concise diagnosis of coverage/quality.",
  "suggestions_if_bad": "If bad: concrete query/strategy improvements for next round."
}
```

Now judge the retrieval-plan quality:
\end{lstlisting}

\subsection{Theorem-Proving Prompts}
\label{appd:prompt:proving}

The prover generates and revises a tactic body within a fixed theorem context.
In both proof-generation prompts, \texttt{\{integrity\_rules\}} contains the shared instructions below.
The \texttt{\{failure\_memory\_guidance\}} field is empty when memory is disabled and contains the instruction in \cref{appd:prompt:memory} when memory is enabled.
The remaining fields supply the immutable source context, retrieved declarations, previous tactic body, and compiler feedback.
\Needspace{9\baselineskip}
\subsubsection{Shared proof-integrity rules.}
\label{appd:prompt:integrity}
\begin{lstlisting}[style=LGAppendixPrompt]
## Non-negotiable Proof-Integrity Rules

These rules are part of the task, not optional style guidance:
1. Prove the supplied target using its hypotheses and existing library results. Do not add an unproved assumption or use the theorem being defined as its own proof. Introducing a binder or hypothesis required by the goal, case analysis, induction, and proof by contradiction are legitimate proof steps.
2. Never use or introduce `axiom`, `constant`, `opaque`, `unsafe`, `unsound`, `sorry`, or `admit`. Do not hide these constructs in another declaration.
3. The host owns every top-level declaration. Generate only the tactic body; no new top-level declarations, imports, namespaces, options, or commands.
4. The supplied prefix, including the target theorem header ending in `:= by`, and the supplied suffix are immutable. The host inserts your generated tactic body between them; you must not copy, rewrite, or close either one.
5. Output only tactic commands that belong inside the existing `by` proof body, without another outer `by`. Local `have`, `let`, `letI`, `haveI`, `suffices`, and ordinary tactic steps are allowed. If you need a helper result, state and prove it locally with `have` or `suffices`, including a stronger or universally quantified helper when useful. Every helper needs a proof. The `classical` tactic and existing library automation are allowed.
6. Existing declarations from the supplied Mathlib environment are the only admissible external facts. Search results are hints to existing declarations, not assumptions and not authorization to invent similarly named facts.
7. Aim to close every goal. If unsuccessful, submit a concrete proof attempt for compiler feedback, without placeholders or trusted-code escape hatches.
The host checks source boundaries and forbidden constructs before compilation; Lean checks the proof. A rejected candidate returns compiler-tool feedback to the next reflection round and does not count as a successful proof.
\end{lstlisting}
\Needspace{9\baselineskip}
\subsubsection{Initial proof attempt.}
\label{appd:prompt:prover_init}
\begin{lstlisting}[style=LGAppendixPrompt]
You are an expert at writing sound Lean 4 proofs. The host will insert exactly one generated tactic body into the proof hole below. Solve the stated theorem; do not reproduce or edit the surrounding Lean source.

## Exact Immutable Prefix

```lean4
{proof_prefix}
```

This prefix already ends in `:= by`. It is context for understanding the goal, not text to repeat in your response.

## Exact Immutable Suffix

```lean4
{proof_suffix}
```

This suffix is host-owned context and must not be repeated or closed by your response.

## Relevant Search Results

{search_results}
Use the search results as hints to declarations that actually exist in the supplied Mathlib environment. You may also use other existing Mathlib declarations you know, but never invent a declaration or add an assumption.
Check a result's type, hypotheses, argument order, and conclusion before using it. Search rank and informal similarity do not establish applicability.
Empty search results do not mean the needed fact is absent from Mathlib.
Any supplied reasoning sketch is a candidate route: adapt it if its steps do not fit the formal target or available declarations.

## Proof Construction

Read the exact target and local hypotheses, then choose a proof that fits them.
Use a direct library application when applicable; otherwise reduce the goal with explicit intermediate claims. Match coercions and typeclass assumptions at each application. Use local `have` blocks for auxiliary results and ordinary Lean tactics for the resulting goals.

{integrity_rules}

{failure_memory_guidance}

## Response Format
First give a concise proof plan. Then output exactly one Lean 4 code block that contains only the honest tactic commands to place after the existing `:= by`.
Return the entire replacement body, including all branches and helper proofs.
Start outermost tactics at column zero; indent nested blocks consistently.
Do not include the theorem statement, imports, namespace commands, suffix, or an additional outer `by`. Put every helper inside the tactic body with `have`, `suffices`, or `let`.
\end{lstlisting}
\Needspace{9\baselineskip}
\subsubsection{Proof revision after feedback.}
\label{appd:prompt:prover_reflect}
\begin{lstlisting}[style=LGAppendixPrompt]
You are an expert at repairing sound Lean 4 proofs. The host owns the immutable source contract and will insert your revised tactic body into its existing `:= by` proof hole.

## Exact Immutable Prefix

```lean4
{proof_prefix}
```

## Exact Immutable Suffix

```lean4
{proof_suffix}
```

## Previous Failed Tactic Body

```lean4
{proof}
```

## Compiler-Tool Feedback

{error_msg}

## Relevant Search Results

{search_results}

Use compiler feedback to choose the next concrete repair. Reported line and column numbers refer to the complete assembled file (prefix, body, suffix), not line numbers within the body shown above.
Prioritize an early syntax or scope error that could cause later errors.
For indentation or parser errors, repair the block structure first. For a type mismatch, compare the expected and actual types, explicit arguments, coercions, and required instances. For remaining goals, prove the displayed obligation. For `no goals to be solved`, remove the redundant tactic.
For an unknown identifier, first determine whether it is a local name, a namespace-resolution issue, or a library declaration. A failed name does not establish that the mathematical fact is absent. Use an applicable known or retrieved declaration, or prove the needed fact locally. Library facts outside the search results remain available. Check retrieved statements against the current goal and hypotheses; a search ranking is not evidence of applicability.
Preserve coherent parts of the proof while repairing the identified obstacle.
Change strategy if the needed step cannot be justified or the chosen approach has stopped making progress. A supplied reasoning sketch is revisable.

{integrity_rules}

{failure_memory_guidance}

## Mandatory Repair of the Previous Attempt

If feedback reports a proof-integrity violation, remove the offending construction. Prove any useful helper locally using `have` or `suffices`, then use it to finish the original target.

## Response Format

First briefly identify the obstacle and the change that addresses it. Then output exactly one Lean 4 code block containing only the repaired tactic commands to place after the existing `:= by`. Return the entire replacement body, not a patch or only the last failing step. Start outermost tactics at column zero and indent nested blocks consistently. Do not repeat the prefix, suffix, theorem header, imports, namespace commands, or an additional outer `by`.
\end{lstlisting}
\Needspace{9\baselineskip}
\subsubsection{Reflection-time retrieval queries.}
\label{appd:prompt:reflect_queries}
\begin{lstlisting}[style=LGAppendixPrompt]
Generate exactly 5 complementary Mathlib search queries for the next proof attempt. These queries supply candidate declarations to a Lean prover; they do not execute Lean code.

## Formal Target and Context
```lean4
{lean_code}
```

## Informal Description
{informal_statement}

## Previous Failed Tactic Body
```lean4
{proof}
```

## Compiler Feedback
{error_msg}

Find the mathematical obligation or missing API behind the current obstacle.
For a library-name error, describe the intended mathematical fact in natural language, including relevant hypotheses and the type of objects. Check whether the unknown name is instead a local variable, a namespace issue, or a result of an earlier syntax error. A local-name or indentation error itself does not require a missing library lemma; search for facts supporting the intended next proof step in that case.
For a type mismatch, seek a version of the fact with the actual domains, codomains, assumptions, and coercions. For an unsolved goal, describe that goal and the hypotheses available to discharge it. Include a definition or bridge lemma when it would connect the available facts to the goal.
Each query should express a distinct useful fact or alternative approach, avoiding repeated paraphrases of one guessed name. Give natural-language descriptions; add an identifier only when it helps disambiguate a known declaration. Explicitly say when searching for a definition. Use only the supplied context, not an assumed benchmark answer. A previous failed application does not make every use of that lemma invalid.
Return exactly one JSON object in a json code block, with no other text:
{"queries": ["natural-language query", "..."]}
The queries array must contain exactly 5 nonempty strings.
\end{lstlisting}

\subsection{Failure-Memory Prompt and Feedback}
\label{appd:prompt:memory}

When memory is enabled, the following instruction is inserted verbatim into the initial proof prompt and every proof-revision prompt.
It applies only to the current proving trajectory and is not part of standalone MathlibMPR retrieval.
\Needspace{9\baselineskip}
\subsubsection{Contextual failure-memory guidance.}
\label{appd:prompt:memory_guidance}
\begin{lstlisting}[style=LGAppendixPrompt]
## Failure Memory

The `failure_memory` feedback records earlier failures in this trajectory.
Treat a listed tactic or declaration as evidence about its failed application, not a global ban on that tool. Corrected arguments, goal shape, instances, or local hypotheses can make a new application valid. An unchanged full candidate with a cached failure is blocked; submit a revised candidate.
When a failure repeats, identify what has changed that could resolve it. Keep useful proof steps; change the local application or choose another strategy when the feedback shows a mathematical obstacle. A repeated error alone does not establish that the whole strategy is impossible. Memory summaries guide repairs; Lean checks the submitted proof.
\end{lstlisting}

\paragraph{Structured feedback for repeated failures.}
Compiler feedback includes a structured \texttt{failure\_memory} object in the \texttt{\{error\_msg\}} field.
It contains the latest state fingerprint and occurrence count, together with summaries of recurrent states and their unsuccessful tactics, declaration applications, and modifications.
Once a state has occurred at least twice, the feedback includes the following directive:
\begin{lstlisting}[style=LGAppendixPrompt]
DEAD_END_REPEATED: this formal state has now appeared at least twice. Do not repeat the same tactic, lemma, declaration, or proof modification. You remain free to change the mathematical strategy, rewrite the whole proof, make a different local repair, or issue a new Mathlib query; the controller will not choose for you.
\end{lstlisting}
This structured feedback is supplied to both proof revision and reflection-time query generation.

\end{document}